\documentclass[letterpaper, 10 pt, conference]{ieeeconf}  

\IEEEoverridecommandlockouts                              

\usepackage{graphicx}
\usepackage{amsmath}
\usepackage{amsfonts}
\usepackage{booktabs}
\usepackage{wrapfig}
\usepackage{tabularx}
\usepackage{subcaption}
\usepackage{multirow}
\usepackage{makecell}
\usepackage[hidelinks]{hyperref}
\newcolumntype{C}{>{\centering\arraybackslash}X}  
\usepackage{xcolor}
\definecolor{berkeleyblue}{HTML}{003262}
\definecolor{magentaPink}{HTML}{C94F9D}

\makeatletter
\newcommand{\independenttablecaption}[2]{%
  \refstepcounter{table}%
  \begingroup
  \centering
  \@makecaption{\fnum@table}{#1}\par
  \endgroup
  \label{#2}%
}
\makeatother

\title{\LARGE \bf Modeling and Understanding Neural Weights of \\ Humanoid Policies with Diffusion
\\ 
}

\author{
Huixuan Cheng, Bike Zhang, Shankar Sastry, and Koushil Sreenath
\thanks{The authors are with the University of California, Berkeley.}
}

\begin{document}

\maketitle
\thispagestyle{empty}
\pagestyle{empty}

\begin{abstract}
Recent advances in robot learning have driven rapid progress in humanoid control, enabling increasingly agile behaviors in the real world. However, humanoid control networks are typically obtained through conventional training procedures, while comparatively little attention has been given to directly modeling the distribution of their learned weights and generating functional networks in weight space. 
In this work, we investigate whether high-performing humanoid control networks can be generatively modeled directly through their weights. We adopt a latent diffusion framework that couples a linear autoencoder with a Diffusion Transformer. The autoencoder compresses actor network parameters into a compact latent space, and the Diffusion Transformer models the distribution of these latents. Trained on networks collected from reinforcement learning for the Unitree G1 robot, our approach generates deployable networks that perform various motions in the real world, including walking, dancing, and spin kick. Beyond generation, we further study the learned weight space to better understand the structure captured by these humanoid control networks. We identify a structural principle in the first-layer weights that consistently reflects the robot's kinematic prior. Notably, this kinematic prior emerges under both Reinforcement Learning (RL) and Imitation Learning (IL) training, suggesting that it is not specific to a particular optimization process but may instead reflect common knowledge associated with humanoid embodiment. Together, these results demonstrate the feasibility of generative modeling in humanoid network weight space, while providing initial insights into the information encoded by their learned weights. Additional results are available on our \href{https://humanoid-weight-diffusion.github.io/}{\textcolor{magentaPink}{project website}}.
\end{abstract}




\section{Introduction}
\label{sec:intro}

\begin{figure}
    \centering
    \includegraphics[width=\linewidth]{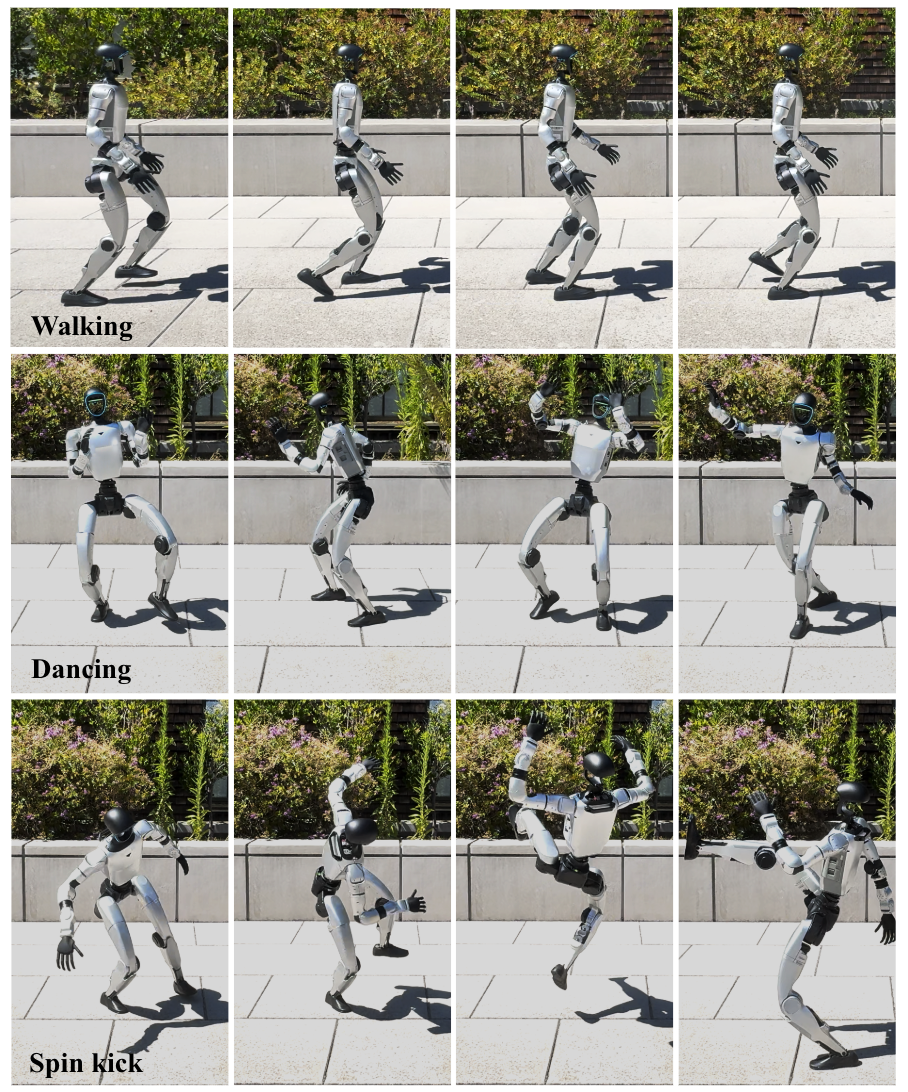}
    \caption{\textbf{Real-world deployment of networks with generated weights.} Top: A walking motion. Middle: A dancing motion. Bottom: A spin kick motion. For each motion, the corresponding network weights are denoised by a motion-specific diffusion model and subsequently deployed zero shot in the real world.}
    \label{fig:motion}
    \vspace{-1em}
\end{figure}

\begin{figure*}[!t]
    \centering
    \includegraphics[width=0.95\textwidth]{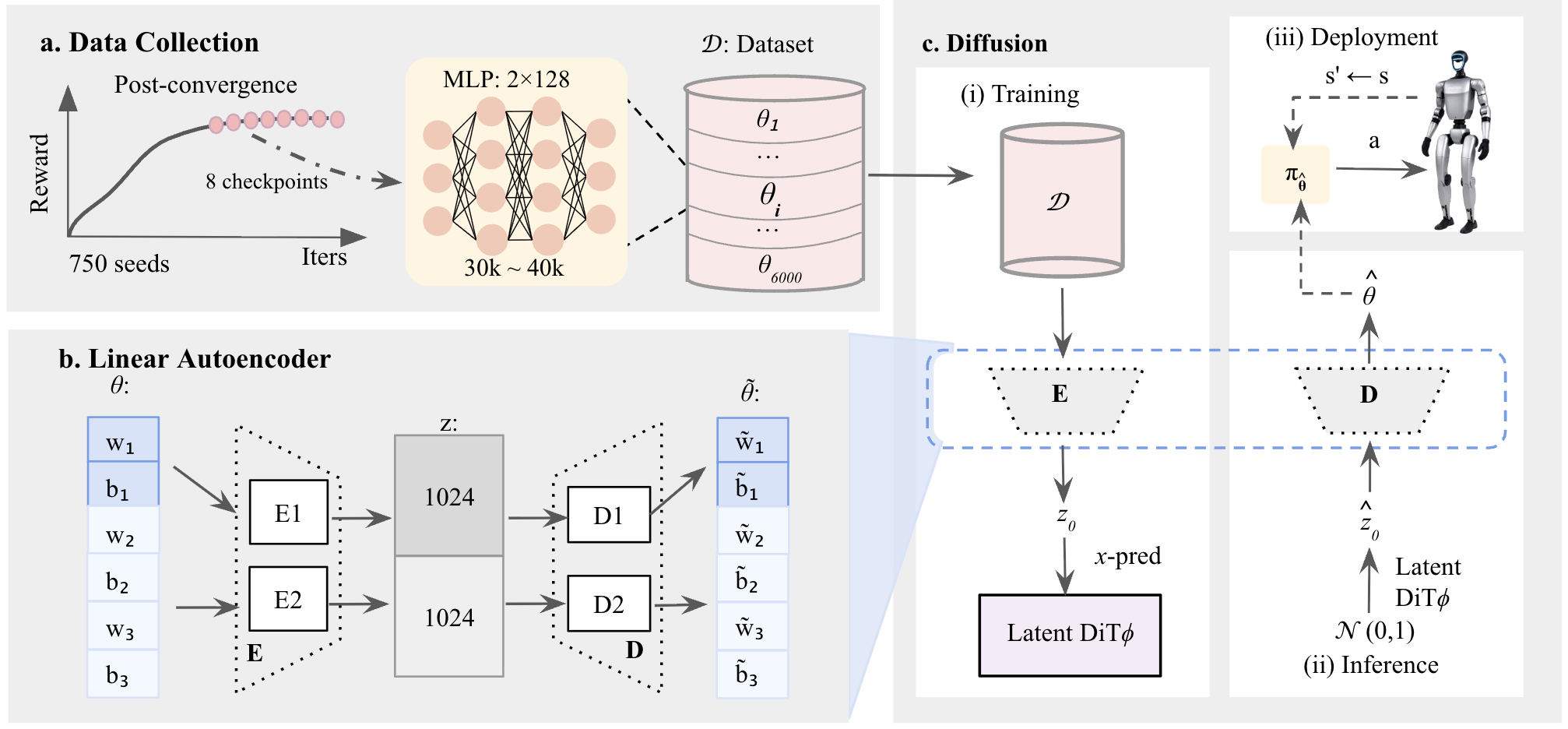}
    \caption{\textbf{Method overview.} (a) We collect actor weights from 750 training runs with random seeds and retain eight post-convergence networks from each run. Each network is a three-layer MLP. (b) The cluster-wise linear autoencoder is implemented using single-layer encoders and decoders. (c) A diffusion transformer is trained with $x$-pred loss to model the latent distribution. During inference, the model generates novel network weights that can be deployed in the real world. }
    \label{fig:overview}
\end{figure*}

Humanoid control has progressed rapidly in recent years, from locomotion to whole-body control, with increasingly agile and versatile motions~\cite{fu2024humanplus, he2024omnih2o, ji2024exbody2}.
Highly dynamic humanoid motions can now be learned through imitation of human reference motions~\cite{liao2025beyondmimic}.
Scaling such motions has further led to humanoid foundation models with broader capabilities~\cite{luo2025sonic}.
Together, these advances enable humanoid robots to interact with complex real-world environments.

Despite this progress, humanoid skills are still typically acquired by training separate control networks for different tasks, which are then maintained as individual expert policies.
An alternative is to directly model and generate these expert networks in weight space.
A generative model over network weights could serve as a policy generator, synthesizing control networks for different motions or tasks, potentially conditioned on task descriptors, reference trajectories, or other embeddings, and providing a compact and flexible representation of diverse humanoid skills.

Early approaches to weight generation, such as hypernetworks, used a learned generator conditioned on a task or context to produce the weights of a target network~\cite{ha2016hypernetworks, sarafian2021recomposing}.
More recent work has explored diffusion-based methods for modeling and generating network weights.
In the image domain, diffusion-generated weights can perform image classification, with recent extensions to language classification and tabular prediction~\cite{wang2024neural, gupta2026deepweightflow, zeng2025generative, peebles2022learninglearngenerativemodels}.
This direction has since extended to embodied control, where latent diffusion models conditioned on target behaviors or trajectories generate network weights for HalfCheetah and Walker in simulation~\cite{hegde2023generating}, with recent progress extending to manipulation tasks~\cite{hegde2025latent, rocca2025policy}.
Notably, weights for quadruped locomotion can also be sampled and deployed in the real world~\cite{liang2024makeagent}.
Generative modeling of humanoid control networks, however, remains largely unexplored, in part due to their high-dimensional weight space.
Neural network weights are known to be substantially redundant, with an effective dimensionality far below the nominal parameter count~\cite{yang2022capacity, huh2021low, transtrum2015sloppy, mao2024training}, suggesting that humanoid network weights may likewise be modeled through compact latent representations.

Beyond weight generation, a complementary question is what information is encoded in humanoid control network weights.
To date, weight space analysis has largely focused on the image domain, where features are comparatively easy to associate with individual layers, from low-level edges and colors to class-specific concepts~\cite{zeiler2014visualizing, yosinski2014transferable}.
In robotics, by contrast, such analysis remains limited: existing studies mostly examine which weights matter, for instance using pruning to identify task-relevant inputs~\cite{vischer2021lottery}.
A systematic understanding of what humanoid control networks encode in their weights therefore remains largely unexplored.
Such analysis could reveal regularities that inform more compact architectures, while structures shared across motions may guide transfer learning for acquiring new policies. 

In this work, we address two gaps: generating humanoid control networks in weight space and interpreting the structure and information encoded in their weights.
We collect a large dataset of RL-trained networks for the Unitree G1 humanoid robot performing walking, dancing, and spin kick motions, and train a linear autoencoder together with a Diffusion Transformer~\cite{peebles2023scalable} to model their weight distribution.
We evaluate the quality and diversity of generated networks in simulation and assess their sim-to-real transfer on hardware, as shown in Fig.~\ref{fig:motion}.
Alongside this generative modeling effort, we systematically analyze the collected humanoid networks, investigating the structure encoded by RL-trained networks and how different training paradigms, including imitation learning, and training configurations, such as domain randomization, affect their weight space distributions.

\begin{table*}[!t]
\centering
\caption{Generated Networks vs. Training Data}
\label{tab:with-dataset}
{\footnotesize
\begin{tabularx}{\textwidth}{l *{8}{C}}
\toprule
 & \multicolumn{2}{c}{Dataset} & \multicolumn{2}{c}{DiT (mean)}
 & \multicolumn{2}{c}{DiT (best 90\%)} & \multicolumn{2}{c}{DiT (best)} \\
\cmidrule(lr){2-3} \cmidrule(lr){4-5} \cmidrule(lr){6-7} \cmidrule(lr){8-9}
Motion & Succ $\uparrow$ & TE $\downarrow$ & Succ $\uparrow$ & TE $\downarrow$
       & Succ $\uparrow$ & TE $\downarrow$ & Succ $\uparrow$ & TE $\downarrow$ \\
\midrule
Walking   & $0.999 \pm 0.001$ & $0.162 \pm 0.003$ & $0.993 \pm 0.023$ & $0.191 \pm 0.020$
          & $0.997 \pm 0.003$ & $0.186 \pm 0.013$ & 1.000 & 0.161 \\
Dancing   & $0.998 \pm 0.004$ & $0.037 \pm 0.001$ & $0.941 \pm 0.213$ & $0.042 \pm 0.007$
          & $0.996 \pm 0.006$ & $0.040 \pm 0.003$ & 1.000 & 0.036 \\
Spin kick & $0.997 \pm 0.004$ & $0.046 \pm 0.002$ & $0.943 \pm 0.289$ & $0.053 \pm 0.007$
          & $0.993 \pm 0.009$ & $0.051 \pm 0.004$ & 1.000 & 0.044 \\
\bottomrule
\addlinespace[2pt]
\multicolumn{9}{@{}p{\textwidth}@{}}{\footnotesize
Dataset, DiT-mean and DiT best 90\% results are reported as mean $\pm$ std.
DiT-best-90\% is the mean over the $90$ best of the $100$ generated samples. DiT best (best of $100$ samples)
corresponds to the per-metric optimum. Samples are ranked separately for each metric. TE denotes tracking error: MPKPE (m) for motion tracking, and mean linear velocity command error (m/s) for walking. For walking, the command includes step changes, so the tracking error also captures transient errors during command transitions. 
}
\end{tabularx}
}
\end{table*}

\begin{table*}[!t]
\centering
\caption{Effect of Transformer Embedding Dimension}
\label{tab:tfm}
{\footnotesize
\begin{tabularx}{\textwidth}{@{} l *{6}{C} @{}}
\toprule
Motion & \multicolumn{2}{c}{embedding=512} & \multicolumn{2}{c}{embedding=768} & \multicolumn{2}{c}{embedding=1024} \\
\cmidrule(lr){2-3} \cmidrule(lr){4-5} \cmidrule(lr){6-7}
 & Prog $\uparrow$ & TE $\downarrow$ & Prog $\uparrow$ & TE $\downarrow$ & Prog $\uparrow$ & TE $\downarrow$ \\
\midrule
Walking   & $0.643 \pm 0.373$ & $0.513 \pm 0.284$ & $0.890 \pm 0.253$ & $0.298 \pm 0.139$ & $\boldsymbol{0.997 \pm 0.010}$ & $\boldsymbol{0.191 \pm 0.020}$ \\
Dancing   & $0.536 \pm 0.231$ & $0.076 \pm 0.012$ & $0.651 \pm 0.243$ & $0.071 \pm 0.012$ & $\boldsymbol{0.965 \pm 0.154}$ & $\boldsymbol{0.042 \pm 0.007}$ \\
Spin kick & $0.296 \pm 0.122$ & $0.082 \pm 0.008$ & $0.873 \pm 0.198$ & $0.067 \pm 0.010$ & $\boldsymbol{0.969 \pm 0.136}$ & $\boldsymbol{0.053 \pm 0.007}$ \\
\bottomrule
\addlinespace[2pt]
\multicolumn{7}{@{}p{\textwidth}@{}}{\footnotesize
All results are reported as mean $\pm$ std over diffusion samples.
Prog is task progress in the range $[0, 1]$: for motion tracking, it is the fraction of reference motion frames completed; for walking, it denotes the fraction of the max episode duration (20 s) survived. TE denotes tracking error: MPKPE (m) for motion tracking, and mean linear velocity command error (m/s) for walking.}
\end{tabularx}}
\end{table*}

\begin{table*}[!t]
  \centering
  \caption{Comparison between diffusion-generated networks and noise-perturbed training networks across motions.}
  \label{tab:noise-injection}
  {\footnotesize
  \begin{tabularx}{\textwidth}{@{} l *{9}{C} @{}}
    \toprule
    & \multicolumn{3}{c}{Walking} & \multicolumn{3}{c}{Dancing} & \multicolumn{3}{c}{Spin kick} \\
    \cmidrule(lr){2-4} \cmidrule(lr){5-7} \cmidrule(lr){8-10}
    Group
      & $\sigma$ & Succ $\uparrow$ & TE $\downarrow$
      & $\sigma$ & Succ $\uparrow$ & TE $\downarrow$
      & $\sigma$ & Succ $\uparrow$ & TE $\downarrow$ \\
    \midrule
    Training
      & $0.00$ & $0.999$ & $0.162$
      & $0.00$ & $0.998$ & $0.037$
      & $0.00$ & $0.997$ & $0.046$ \\
    Noise $\sigma{=}0.01$
      & $0.01$ & $0.999$ & $0.161$
      & $0.01$ & $0.997$ & $0.036$
      & $0.01$ & $0.997$ & $0.047$ \\
    Noise $\sigma{=}0.05$
      & $0.05$ & $0.998$ & $0.172$
      & $0.05$ & $0.990$ & $0.042$
      & $0.05$ & $0.984$ & $0.052$ \\
    Noise $\sigma{=}0.10$
      & $0.10$ & $0.959$ & $0.226$
      & $0.10$ & $0.629$ & $0.057$
      & $0.10$ & $0.650$ & $0.068$ \\
    \midrule
    Noise, matched
      & $0.110$ & $0.914$ & $0.248$
      & $0.078$ & $0.865$ & $0.051$
      & $0.090$ & $0.783$ & $0.064$ \\
    Diffusion
      & $0.110$ & $\mathbf{0.993}$ & $\mathbf{0.191}$
      & $0.078$ & $\mathbf{0.941}$ & $\mathbf{0.042}$
      & $0.090$ & $\mathbf{0.943}$ & $\mathbf{0.053}$ \\
    \bottomrule
    \addlinespace[2pt]
    \multicolumn{10}{@{}p{\textwidth}@{}}{\footnotesize
    All results are reported as mean over 100 networks. Distance to the training distribution is expressed as an equivalent noise level $\sigma$. The upper block reports performance under a fixed noise schedule applied to training networks. The lower block reports a distance-matched baseline, where the noise level is set to match the equivalent distance of the diffusion-generated networks, alongside the diffusion results. Succ and TE are the same as in Table \ref{tab:with-dataset}. }
  \end{tabularx}}
\end{table*}

\section{Methodology}
\label{sec:method}

In this section, we will present our method to generatively model policy weights, as illustrated in Fig.~\ref{fig:overview}. We first describe the collection process for the weight dataset used for generative model training, and then introduce our approach, which consists of two components: (1) a linear autoencoder that compresses actor network weights into compact latent representations, and (2) a diffusion transformer trained on latent space to model their distribution.

\noindent\textbf{Weight Dataset.}
We collect a dataset $\mathcal{D}$ of actor networks for the Unitree G1 robot across three motions: walking, dancing, and spin kick. Each motion is trained via RL in mjlab~\cite{mjlab}, and we retain 6{,}000 actor networks per motion. 
All actor networks share a common architecture: a three-layer multi-layer perceptron (MLP) with hidden dimensions $[128, 128]$ and ELU activations, resulting in approximately 30k $\sim$ 40k parameters per network, with the variation arising from different observation space dimensions for each motion.
To ensure broad coverage of the weight distribution, we launch 750 independent training runs with random seeds and collect 8 post-convergence networks per seed, as shown in Fig.~\ref{fig:overview}a.  We disabled observation normalization during RL training and only actor weights are considered for generation.

\noindent\textbf{Linear Autoencoder.}
\label{sec:ae}
We use a linear autoencoder to compress the actor network weights into latents to reduce dimensionality. 
The encoder and decoder are implemented as single linear layers without bias terms or activation functions. Given an actor network $\pi_\theta$ parameterized by $\theta = \begin{bmatrix}w_1, b_1, w_2, b_2, w_3, b_3\end{bmatrix}\in \mathcal{D}$, we flatten and concatenate the weights $w_i$ and biases $b_i$ of all layers in sequence. The encoder $\mathbf{E}$ is composed of $\mathbf{E}_1$ that compresses the cluster $[w_1, b_1]$ into a 1024-dimensional latent vector and $\mathbf{E}_2$ that compresses the cluster $[w_2,b_2,w_3,b_3]$ into another 1024-dimensional latent vector, resulting in a cluster-wise encoding scheme for $\mathbf{E}$. The two latent vectors are then concatenated to form a 2048-dimensional latent representation $z$ used for diffusion training. The decoder $\mathbf{D}$ reconstructs the network parameters in the same cluster-wise manner, as illustrated in Fig.~\ref{fig:overview}b.
We train separate autoencoders for different motions. The training objective is the mean squared reconstruction error:
$\mathcal{L}_{\mathrm{AE}} = \| \theta - \tilde{\theta} \|_2^2$, where $\tilde{\theta} = \mathbf{D}(\mathbf{E}(\theta)), \theta\in\mathcal{D}$, and $\tilde{\theta}$ consists of the reconstructed weights $\tilde{w}_i$ and biases $\tilde{b}_i$.

\noindent\textbf{Diffusion Transformer.}
Following the latent diffusion paradigm~\cite{rombach2022high}, we train a diffusion transformer (DiT) to model the latent distribution for each motion~\cite{peebles2023scalable, vaswani2017attention}, while keeping the autoencoders frozen. The overall pipeline is detailed in Fig.~\ref{fig:overview}c.
We first encode the actor weight $\theta$ into a clean latent $z_0 = \mathbf{E}(\theta)$, and then inject noise at each timestep $t \in \{1, \dots, T\}$ to produce: 
\begin{equation}
    z_t = \sqrt{\bar{\alpha}_t}\, z_0 + \sqrt{1 - \bar{\alpha}_t}\, \epsilon,
    \quad \epsilon \sim \mathcal{N}(0, I).
    \label{eq:forward}
\end{equation}
where $\bar{\alpha}_t$ is a pre-defined noise schedule. Instead of the conventional $\epsilon$-prediction, we optimize the diffusion model via \textit{x}-prediction, i.e., train the DiT to directly predict the clean latent $z_0$~\cite{x-pred}:
\begin{equation}
    \mathcal{L}_{\mathrm{DiT}}(\phi) = \mathbb{E}_{t,\, z_0,\, \epsilon\sim \mathcal{N}(0,I)}
    \!\left[\left\| \mathrm{DiT}_\phi(z_t, t) - z_0 \right\|^2\right],
    \label{eq:loss}
\end{equation}
where $\mathrm{DiT}_\phi$ is the \textit{x}-pred network parameterized by $\phi$. 
We use a transformer backbone to capture the cross-layer and global correlations in weight space, in contrast to the U-Net backbone in \cite{liang2024makeagent}. 
During inference, $\hat{z}_0$ is denoised from $\mathrm{DiT}_\phi$, and subsequently decoded back to weight space via $\hat{\theta} = \mathbf{D}(\hat{z}_0)$. The generated weight $\hat{\theta}$ then produces the generated actor network $\pi_{\hat{\theta}}$, which can be deployed for sim-to-real transfer.

\begin{figure}
    \centering
    \includegraphics[width=\columnwidth]{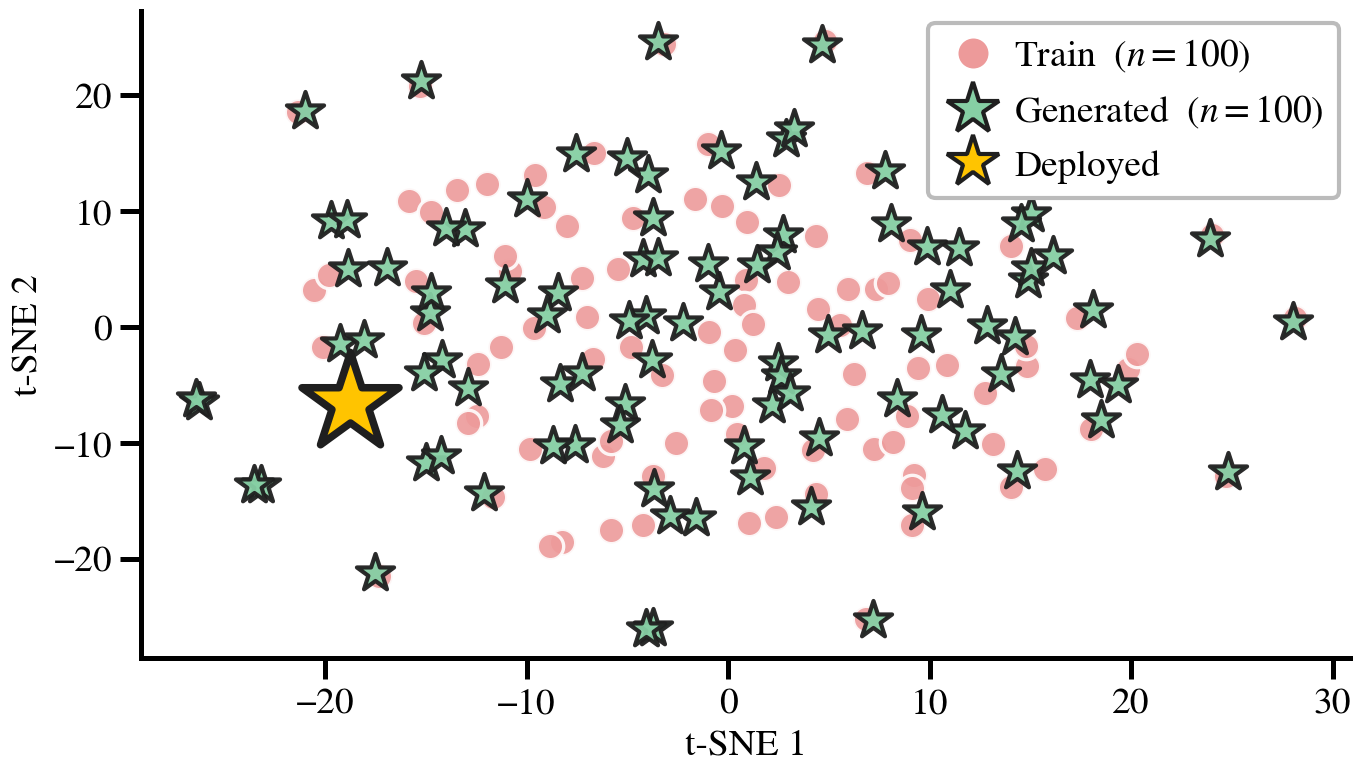}
    \caption{\textbf{t-SNE visualization}. Training and generated
             network weights for the spin kick motion.}
    \label{fig:tsne-generated}
\end{figure}

 \begin{figure*}[!t]
    \centering
    \includegraphics[width=0.9\linewidth, trim=6.4cm 2.2cm 8.1cm 0.7cm, clip]{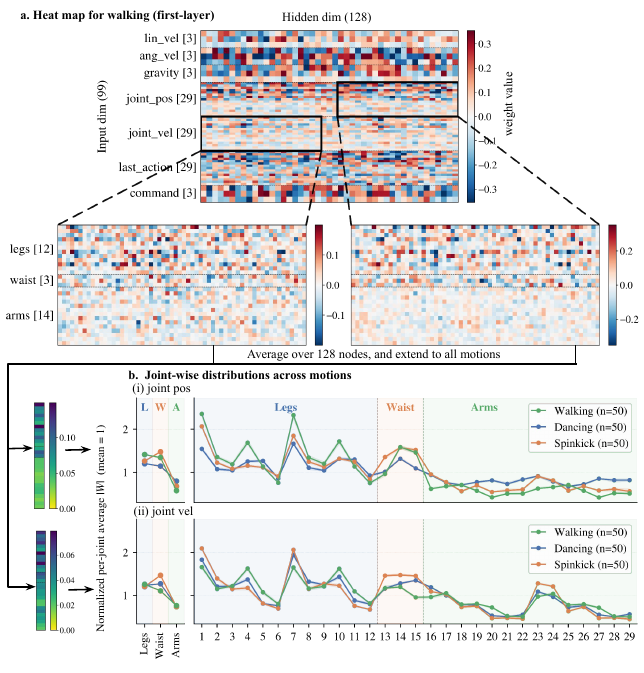}
    \caption{\textbf{Kinematic prior in weights}. (a) A visualization of the first-layer weight heat map for the walking motion, which reveals a structural principle that certain input regions exhibit larger weight magnitudes.
    (b) Joint-wise weight distributions across motions demonstrate that hip pitch, knee, and waist joints are assigned higher importance, suggesting the presence of a kinematic prior.}
    \label{fig:weight-region}
    \vspace{-1em}
\end{figure*}


\section{Weight Generation Experiments}
\label{sec:experiments}

After training the autoencoder and the diffusion transformer, we evaluate the proposed approach in this section. Our experiments are designed to assess: (1) the motion performance of the generated actor network weights, measured by success rate and tracking error, compared to baselines, (2) the sim-to-real transfer capability on Unitree G1, and (3) whether the generated weights merely exhibit memorization of the training network weights or generalization out of the training distribution.

\subsection{Experimental Setup}
\label{sec:exp-setup}

\noindent\textbf{Simulation Environment.}
We use mjlab~\cite{mjlab} as the simulation environment for all motions.
We consider three motions of increasing complexity: (T1) Walking, basic locomotion; (T2) Dancing, a rhythmic whole-body tracking motion; and (T3) Spin kick, a highly dynamic kicking motion.

\noindent\textbf{Baselines.}
We adopt two sets of baselines: (1) the RL-collected network dataset, and (2) networks generated by diffusion transformers with different embedding dimensions, namely 512, 768, and 1024 (ours). Performance is evaluated using success rates (Succ) and tracking errors (TE) in simulation environment.

\subsection{Results}
\label{sec:results}

\noindent\textbf{Motion Quality.}
We first compare the generated networks with the RL-collected networks in terms of success rate and tracking error across three motions. The tracking error is defined as the mean linear velocity error for walking and mean per-keybody position error (MPKPE) for dancing and spin kick.
The success rate is computed as the fraction of successfully completed episodes across 1{,}024 parallel environments.
As shown in Table~\ref{tab:with-dataset}, DiT-generated networks achieve slightly lower average performance than the dataset networks, while the best of 100 generated samples reaches comparable performance. To further assess generation quality, we report the success rate and tracking error over the top $90\%$ of generated samples. Their mean and variance are comparable to those of the RL-collected networks, indicating that only fewer than $10\%$ of generated samples exhibit performance collapse.

We ablate the transformer embedding dimension, which we find to be a critical factor affecting performance during the training process.
We measure the task progress (Prog) and tracking error (TE) across different motions and models, as detailed in Table~\ref{tab:tfm}. Task progress is defined as the fraction of reference motion frames completed for dancing and spin kick, and as the fraction of the maximum episode duration (20 s) survived for walking. 
We observe a clear improvement in performance as the embedding dimension increases, which is most pronounced for the spin kick motion, followed by dancing and walking.
This observation suggests that highly dynamic motions are associated with more complex weight distributions, thereby requiring higher-capacity models to capture them effectively.

\noindent\textbf{Sim-to-Real Transfer.}
To validate the practical applicability of our approach, we deploy the generated actor networks on the Unitree G1 robot without additional fine-tuning.
Since the RL-collected networks are trained with domain randomization, the sim-to-real gap is implicitly incorporated into the weight distribution, eliminating the need for additional sim-to-real considerations during diffusion training.
Once the model is trained, we can sample new networks and directly apply the same deployment protocol in real-world experiments~\cite{liao2025beyondmimic}.
The deployed networks successfully perform all motions, as shown in Fig.~\ref{fig:motion}.

\noindent\textbf{Memorization vs.\ Generalization.}
A natural question is whether our approach merely memorizes the training weights or instead learns the underlying weight distribution. To investigate this, we embed both the training and generated weights into a shared t-SNE space using 100 samples from each set, as shown in Fig.~\ref{fig:tsne-generated}. Taking spin kick as an example, the generated samples (stars) largely lie within the support of the training distribution (dots), indicating that our approach primarily produces in-distribution weights.
We find that even samples appearing at the boundary of the training distribution can successfully perform real-world spin kick on G1.

To examine whether the diffusion model generalizes beyond memorizing training weights or their local noisy variants, we compare diffusion-generated networks with noise-perturbed training networks using the previously defined success rate and tracking error. We define the noise level $\sigma$ as a normalized distance in weight space. For each weight matrix or bias vector $W_l$, we apply
\begin{equation*}
W_l' = W_l + \sigma s_l \varepsilon_l,\quad
\varepsilon_l\sim\mathcal{N}(0,I),\quad
s_l=\operatorname{Std}(W_l),
\end{equation*}
where $W_l'$ denotes the perturbed weights, $s_l$ is the standard deviation of $W_l$, $\varepsilon_l$ is i.i.d. Gaussian noise, and $\sigma$ controls the perturbation magnitude relative to the scale of $W_l$.

For a diffusion-generated network $D$, we identify its nearest training network $W$ in weight space and define the equivalent noise level as the root mean square of $(D_l-W_l)/s_l$ over all weights and biases, where $D_l$ and $W_l$ denote the corresponding parameters of the generated and training networks. We then perturb training networks using this matched noise level and the same layer-wise scaling, yielding a distance-matched baseline for comparison with diffusion-generated networks. The simulation results for all motions are reported in Table~\ref{tab:noise-injection}.

As shown, at matched distance from the training set, diffusion-generated networks consistently outperform noise-perturbed baselines and achieve higher success rate and  lower tracking error across all motions. These results indicate that the diffusion model captures a structured weight-space distribution that supports valid, high-performing solutions beyond individual training samples. We characterize this behavior as local generalization, where the model learns the local geometry of the weight manifold and generates functionally valid neighboring networks.



\begin{figure*}[t]
    \centering
    \includegraphics[width=0.9\linewidth]{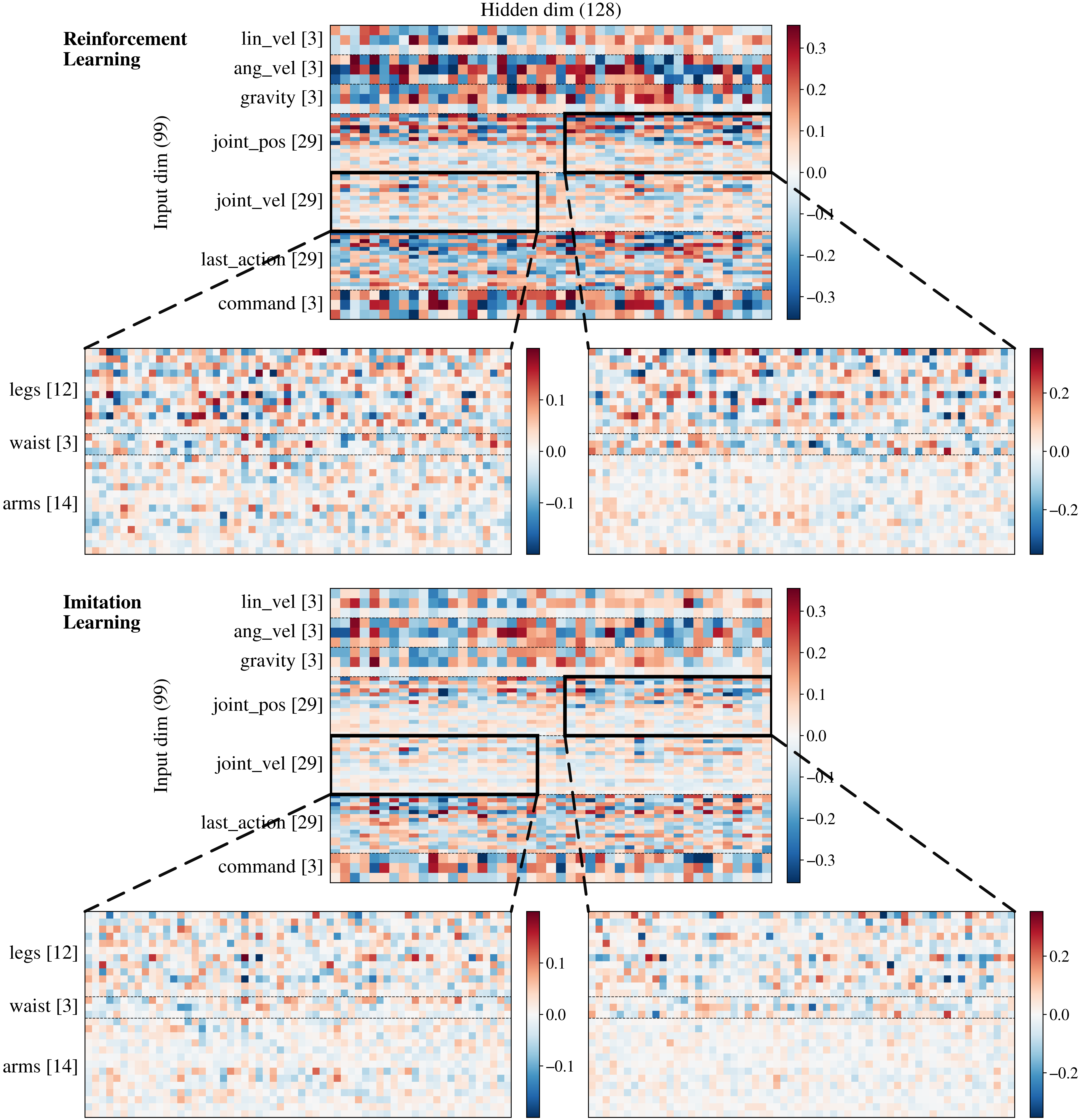}
    \caption{\textbf{First-layer weight comparison between RL and IL for walking}. Top: RL-trained weight heat map. Bottom: IL-trained weight heat map, distilled from the top RL network. While trained via different optimization objectives, they both exhibit kinematic prior in the first-layer weights. }
    \label{fig:bcrl}
    \vspace{-1em}
\end{figure*}


\section{Weight Space Analysis}
\label{sec:weight-analysis}
Having presented the experimental results of the network weight denoising approach, we next provide an analysis of the weight space.
Our study primarily focuses on the first layer of RL-trained actor networks, whose inputs carry direct semantic meaning.
We apply observation normalization to reduce the effect of different input magnitudes.
This analysis aims to identify structural patterns in the learned weights and examine how consistently they emerge across training settings. 
Specifically, we investigate:
(i) how weights behave across different input regions and motions,
(ii) 
how weights differ between RL and IL for the same motion
and (iii) how domain randomization during RL training affects the weight-space distribution.



\subsection{Kinematic Prior in Weights}
\label{sec:sp}
We visualize the first-layer weight matrix in Fig.~\ref{fig:weight-region}a, where darker colors indicate larger weight magnitudes. Building upon prior analysis of RL network weights suggesting that parameters with larger magnitudes tend to be more important~\cite{vischer2021lottery}, we observe that certain input regions, such as \texttt{joint pos} and \texttt{gravity}, consistently exhibit larger magnitudes than others, including \texttt{last action} and \texttt{command}. The complete observation space is illustrated in the left column of Fig.~\ref{fig:weight-region}a.

We further analyze the inputs shared across motions, namely \texttt{joint pos} and \texttt{joint vel}, each consisting of 29 dimensions corresponding to all DoFs, ordered as legs (12), waist (3), and arms (14). For each joint, we compute the mean absolute weight value across the 128 hidden nodes and further average it over 50 random seeds, yielding a measure of per-joint importance.
Although the overall weight magnitudes vary across motions, the normalized joint-wise distributions are highly consistent, as illustrated in Fig.~\ref{fig:weight-region}b. This observation reveals a structural principle in the first-layer weights: the network assigns the greatest importance to the waist region and to joint indices 1 and 7, corresponding to the left and right hip pitch joints, which collectively define the pelvic region. These are followed by the remaining joints in legs and then the arms. We therefore hypothesize that the first-layer weight matrix captures a \emph{kinematic prior} rather than motion-specific knowledge. This interpretation aligns with findings in human motor control and showing that the trunk primarily provides postural stability, while the limbs are responsible for mobility and motion execution~\cite{winter2009biomechanics}.



\begin{figure}
    \centering
    \includegraphics[width=\columnwidth]{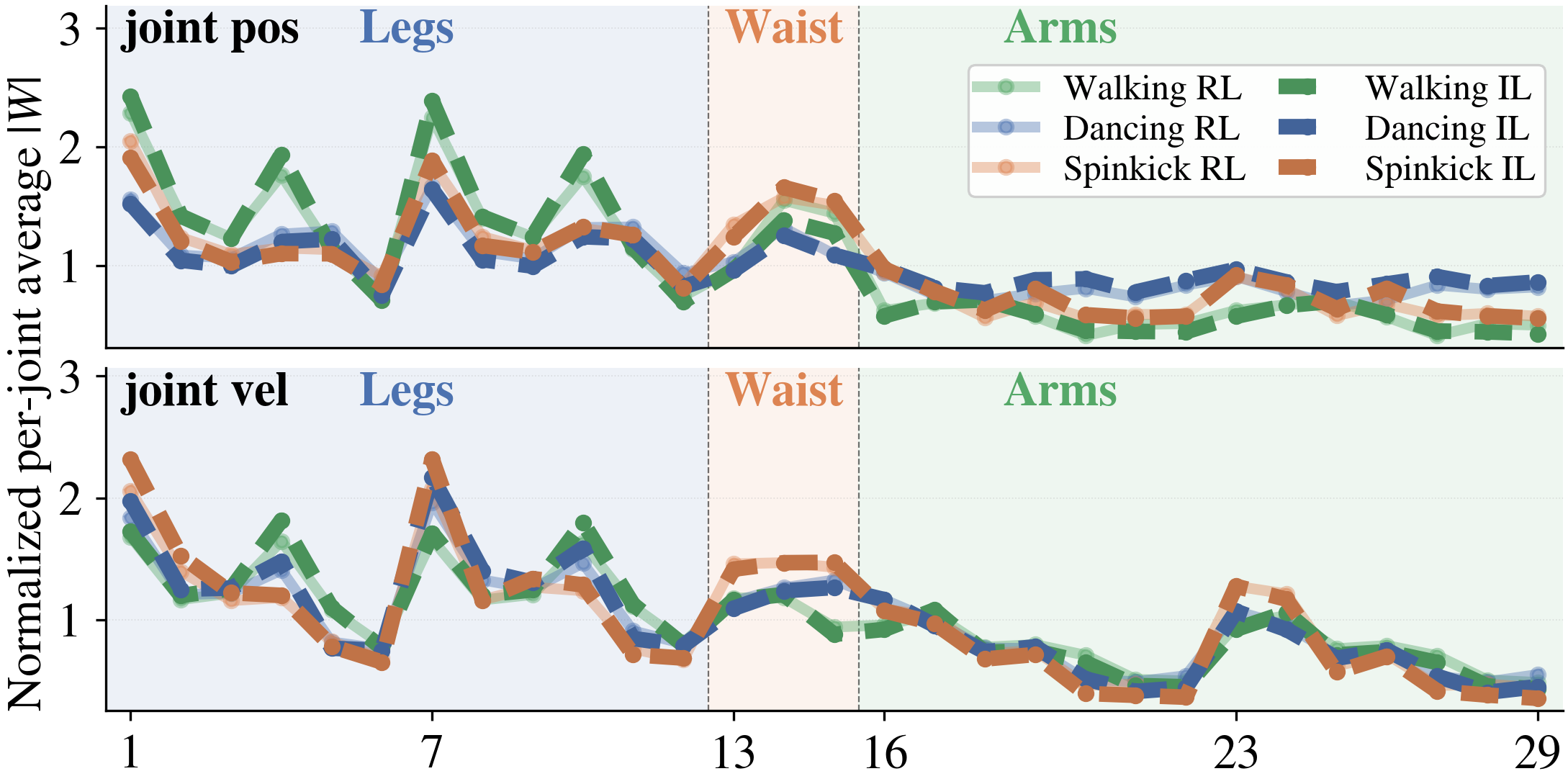}
    \caption{\textbf{Kinematic prior for both RL and IL networks.} For each motion, the dashed and solid curves largely overlap after distillation, indicating that the first-layer kinematic prior is preserved in the IL network. }
    \label{fig:rlil_prior}
    \vspace{-1em}
\end{figure}

\subsection{RL and IL Weight Comparison}
The emergence of a kinematic prior in the weight space raises a natural question: does it depend on how the network is trained? To investigate whether the same kinematic prior persists across different training paradigms, we further analyze networks trained via imitation learning (IL), or policy distillation, to perform the same motions.

For each motion, we distill an IL network from a pretrained RL network. We rollout the RL network in parallel environments to collect a demonstration buffer, with each trajectory covering a complete motion episode. The resulting state-action pairs are then used to train the IL network via offline supervised regression with MSE loss. The IL network uses the same initialization, architecture, activation functions, observation normalization, and inputs as the RL network. The distilled network successfully completes the target motion, yielding a functionally equivalent network obtained through an entirely different optimization process. We distill 50 RL networks for each motion and obtain 50 IL networks respectively. The simulation benchmark of success rate and tracking error is provided in Table~\ref{tab:functional_equivalence}.

Fig.~\ref{fig:bcrl} compares the first-layer weight heatmaps of RL and IL networks for walking. Despite being trained through different optimization processes, the RL and IL networks exhibit a similar kinematic prior in their first-layer weights. As shown in Fig.~\ref{fig:rlil_prior}, when averaged over 50 networks for each training paradigm and motion, the normalized per-joint importance curves of RL and IL nearly overlap across all motions. Nevertheless, the resulting RL and IL solutions are not trivially connected in weight space: RL weights tend to have larger magnitudes, and interpolation between RL and IL does not yield a network that can complete the motion. This suggests that the observed kinematic prior is not specific to the RL optimization process, but may instead reflect common knowledge associated with humanoid embodiment and the underlying motion kinematics. Such training method independent structure further suggests that the kinematic prior could potentially serve as an initialization or warm start for future humanoid policy learning. 
\begin{table}[t]
\centering
\caption{Simulation results for RL teachers and their distilled IL students.}
\label{tab:functional_equivalence}
\renewcommand{\arraystretch}{1.15}
\footnotesize
\begin{tabular*}{\columnwidth}{@{\extracolsep{\fill}}llll@{}}
\toprule
RL / IL & Walking & Dancing & Spin kick \\
\midrule
Succ & 0.9994\,/\,0.9979 & 0.9973\,/\,0.9948 & 0.9972\,/\,0.9919 \\
TE   & 0.1599\,/\,0.1609 & 0.0340\,/\,0.0348 & 0.0466\,/\,0.0478 \\
\bottomrule
\end{tabular*}
\\[2pt]
\begin{minipage}{\columnwidth}
\footnotesize
All results are reported as the mean over 50 RL or IL networks,
evaluated through 512 parallel rollouts in simulation.
\end{minipage}
\vspace{-1em}
\end{table}



\subsection{Effect of Domain Randomization on Weight Space}
\begin{figure*}
    \centering
    \includegraphics[width=0.90\linewidth]{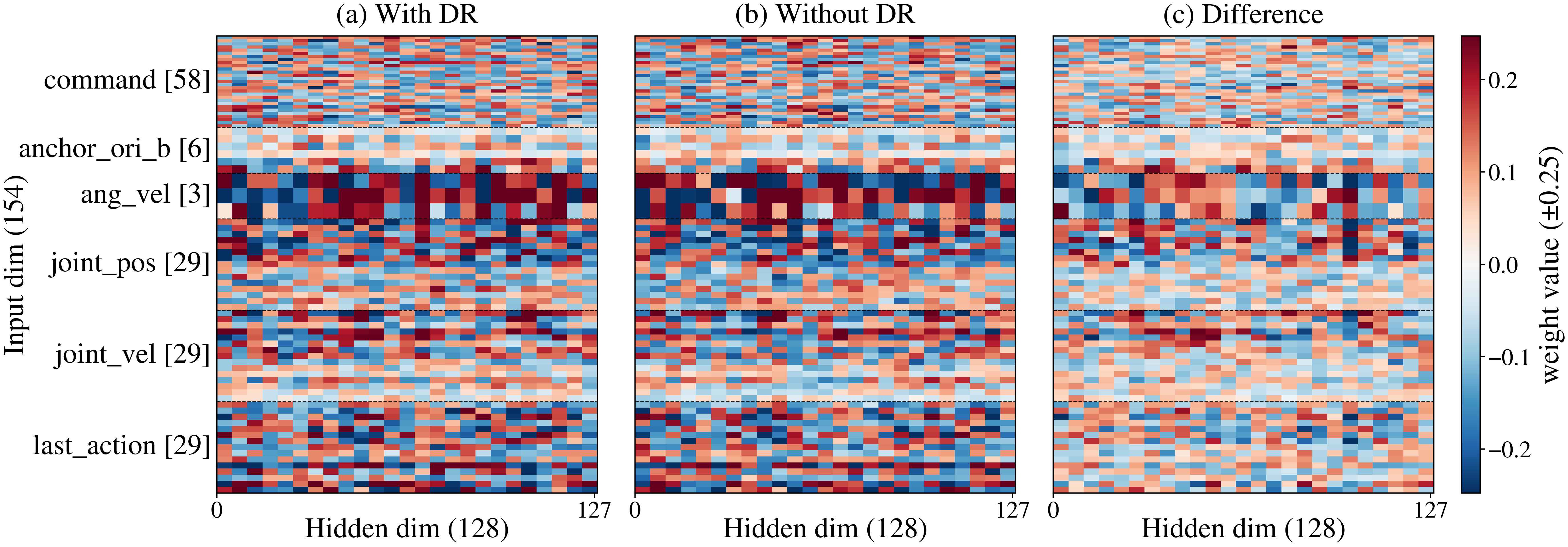}
    \caption{\textbf{Effect of domain randomization on first-layer weights for dancing.} First-layer weight heat maps of RL networks trained from the same initialization (a) with DR and (b) without DR, together with (c) their element-wise difference. Despite their overall similarity, the structured weight difference resembles the RL-trained weights. 
    }
    \label{fig:dr}
    \vspace{-1em}
\end{figure*}

Domain randomization (DR) is a key component of sim-to-real learning, but its effect is typically evaluated only through performance. We instead examine how it changes the weight space. For each motion, we train two RL networks from the same initialization, with and without DR, and visualize their first-layer weight heatmaps and their element-wise weight difference, as shown in Fig.~\ref{fig:dr}.

Although the weights learned with and without DR exhibit highly similar overall structures, their element-wise difference is substantial and itself forms a structure resembling the RL-trained weights. This suggests that DR significantly reshapes the learned weights while preserving their overall distribution. However, the difference weights alone do not constitute a network that can accomplish the motion, nor does linear interpolation between the networks trained with and without DR work across all motions.

These results suggest that the effect of DR cannot be represented as a simple linear displacement in weight space. Instead, DR induces a structured reorganization of the weights throughout RL training while largely preserving the overall weight-space structure associated with the motion.

\section{Limitations and Future Work}
Our analysis is mainly limited to the first-layer, while deeper layers and the effect of domain randomization on weight space remain less understood. The current study is also restricted to a limited set of motions and a single embodiment, and the generality of the observed weight space structures remains to be validated. Although we identify interpretable weight structures, these insights have not yet translated into improved performance or  policy generation.

Future work may extend the analysis to deeper layers and explore conditional weight generation guided by behaviors, trajectories, or text, supported by more diverse datasets across motions and embodiments.


\section{Conclusion}
\label{sec:conclusion}
In this work, we studied humanoid control through the lens of neural network weight space distribution. We showed that latent diffusion can generate functional networks that transfer zero-shot to the real world.  Moreover, our analysis revealed that the first-layer weights consistently encode the robot’s kinematic prior across all motions. We further found that though RL and IL networks are trained through different optimization objectives, they exhibit similar weight space structure. Together, these results provide new insights into the structure and interpretability of humanoid network weights.



\section*{ACKNOWLEDGMENT}
This work was supported in part by the DARPA program Design of Robustly Implementable Autonomous and Intelligent Machines under award HR00112490425. We thank Prof. Claire Tomlin for valuable discussions and insightful feedback, and Qiayuan Liao for generous assistance with the experiments. We are also grateful to Yiyang, Haoyi, Chuye, Gechen, Siming, and Sangli for experimental support. 


\bibliographystyle{IEEEtran}
\bibliography{reference}

\begin{thebibliography}{10}
\providecommand{\url}[1]{#1}
\csname url@samestyle\endcsname
\providecommand{\newblock}{\relax}
\providecommand{\bibinfo}[2]{#2}
\providecommand{\BIBentrySTDinterwordspacing}{\spaceskip=0pt\relax}
\providecommand{\BIBentryALTinterwordstretchfactor}{4}
\providecommand{\BIBentryALTinterwordspacing}{\spaceskip=\fontdimen2\font plus
\BIBentryALTinterwordstretchfactor\fontdimen3\font minus \fontdimen4\font\relax}
\providecommand{\BIBforeignlanguage}[2]{{%
\expandafter\ifx\csname l@#1\endcsname\relax
\typeout{** WARNING: IEEEtran.bst: No hyphenation pattern has been}%
\typeout{** loaded for the language `#1'. Using the pattern for}%
\typeout{** the default language instead.}%
\else
\language=\csname l@#1\endcsname
\fi
#2}}
\providecommand{\BIBdecl}{\relax}
\BIBdecl

\bibitem{fu2024humanplus}
Z.~Fu, Q.~Zhao, Q.~Wu, G.~Wetzstein, and C.~Finn, ``Humanplus: Humanoid shadowing and imitation from humans,'' \emph{arXiv preprint arXiv:2406.10454}, 2024.

\bibitem{he2024omnih2o}
T.~He, Z.~Luo, X.~He, W.~Xiao, C.~Zhang, W.~Zhang, K.~Kitani, C.~Liu, and G.~Shi, ``Omnih2o: Universal and dexterous human-to-humanoid whole-body teleoperation and learning,'' \emph{arXiv preprint arXiv:2406.08858}, 2024.

\bibitem{ji2024exbody2}
M.~Ji, X.~Peng, F.~Liu, J.~Li, G.~Yang, X.~Cheng, and X.~Wang, ``Exbody2: Advanced expressive humanoid whole-body control,'' \emph{arXiv preprint arXiv:2412.13196}, 2024.

\bibitem{liao2025beyondmimic}
Q.~Liao, T.~E. Truong, X.~Huang, Y.~Gao, G.~Tevet, K.~Sreenath, and C.~K. Liu, ``Beyondmimic: From motion tracking to versatile humanoid control via guided diffusion,'' \emph{arXiv preprint arXiv:2508.08241}, 2025.

\bibitem{luo2025sonic}
Z.~Luo, Y.~Yuan, T.~Wang, C.~Li, S.~Chen, F.~Castaneda, Z.-A. Cao, J.~Li, D.~Minor, Q.~Ben \emph{et~al.}, ``Sonic: Supersizing motion tracking for natural humanoid whole-body control,'' \emph{arXiv preprint arXiv:2511.07820}, 2025.

\bibitem{ha2016hypernetworks}
D.~Ha, A.~Dai, and Q.~V. Le, ``Hypernetworks,'' \emph{arXiv preprint arXiv:1609.09106}, 2016.

\bibitem{sarafian2021recomposing}
E.~Sarafian, S.~Keynan, and S.~Kraus, ``Recomposing the reinforcement learning building blocks with hypernetworks,'' in \emph{International Conference on Machine Learning}.\hskip 1em plus 0.5em minus 0.4em\relax Pmlr, 2021, pp. 9301--9312.

\bibitem{wang2024neural}
K.~Wang, D.~Tang, B.~Zeng, Y.~Yin, Z.~Xu, Y.~Zhou, Z.~Zang, T.~Darrell, Z.~Liu, and Y.~You, ``Neural network diffusion,'' \emph{arXiv preprint arXiv:2402.13144}, 2024.

\bibitem{gupta2026deepweightflow}
S.~Gupta, S.~Biggs, M.~Laber, Z.~Shafi, R.~Walters, and A.~Paul, ``Deepweightflow: Re-basined flow matching for generating neural network weights,'' \emph{arXiv preprint arXiv:2601.05052}, 2026.

\bibitem{zeng2025generative}
B.~Zeng, Y.~Yin, Z.~Xu, and Z.~Liu, ``Generative modeling of weights: Generalization or memorization?'' \emph{arXiv preprint arXiv:2506.07998}, 2025.

\bibitem{peebles2022learninglearngenerativemodels}
\BIBentryALTinterwordspacing
W.~Peebles, I.~Radosavovic, T.~Brooks, A.~A. Efros, and J.~Malik, ``Learning to learn with generative models of neural network checkpoints,'' 2022. [Online]. Available: \url{https://arxiv.org/abs/2209.12892}
\BIBentrySTDinterwordspacing

\bibitem{hegde2023generating}
\BIBentryALTinterwordspacing
S.~Hegde, S.~Batra, K.~R. Zentner, and G.~S. Sukhatme, ``Generating behaviorally diverse policies with latent diffusion models,'' 2023. [Online]. Available: \url{https://arxiv.org/abs/2305.18738}
\BIBentrySTDinterwordspacing

\bibitem{hegde2025latent}
\BIBentryALTinterwordspacing
S.~Hegde, S.~Das, G.~Salhotra, and G.~S. Sukhatme, ``Warpd: World model assisted reactive policy diffusion,'' 2025. [Online]. Available: \url{https://arxiv.org/abs/2410.14040}
\BIBentrySTDinterwordspacing

\bibitem{rocca2025policy}
M.~Rocca, S.~Darkner, K.~Erleben, and S.~Andrews, ``Policy-space diffusion for physics-based character animation,'' \emph{ACM Transactions on Graphics}, vol.~44, no.~3, pp. 1--18, 2025.

\bibitem{liang2024makeagent}
Y.~Liang, T.~Xu, K.~Hu, G.~Jiang, F.~Huang, and H.~Xu, ``Make-an-agent: A generalizable policy network generator with behavior-prompted diffusion,'' \emph{Advances in Neural Information Processing Systems}, vol.~37, pp. 19\,288--19\,306, 2024.

\bibitem{yang2022capacity}
\BIBentryALTinterwordspacing
R.~Yang, J.~Mao, and P.~Chaudhari, ``Does the data induce capacity control in deep learning?'' 2022. [Online]. Available: \url{https://arxiv.org/abs/2110.14163}
\BIBentrySTDinterwordspacing

\bibitem{huh2021low}
\BIBentryALTinterwordspacing
M.~Huh, H.~Mobahi, R.~Zhang, B.~Cheung, P.~Agrawal, and P.~Isola, ``The low-rank simplicity bias in deep networks,'' 2023. [Online]. Available: \url{https://arxiv.org/abs/2103.10427}
\BIBentrySTDinterwordspacing

\bibitem{transtrum2015sloppy}
\BIBentryALTinterwordspacing
M.~K. Transtrum, B.~Machta, K.~Brown, B.~C. Daniels, C.~R. Myers, and J.~P. Sethna, ``Sloppiness and emergent theories in physics, biology, and beyond,'' 2015. [Online]. Available: \url{https://arxiv.org/abs/1501.07668}
\BIBentrySTDinterwordspacing

\bibitem{mao2024training}
J.~Mao, I.~Griniasty, H.~K. Teoh, R.~Ramesh, R.~Yang, M.~K. Transtrum, J.~P. Sethna, and P.~Chaudhari, ``The training process of many deep networks explores the same low-dimensional manifold,'' \emph{Proceedings of the National Academy of Sciences}, vol. 121, no.~12, p. e2310002121, 2024.

\bibitem{zeiler2014visualizing}
M.~D. Zeiler and R.~Fergus, ``Visualizing and understanding convolutional networks,'' in \emph{European conference on computer vision}.\hskip 1em plus 0.5em minus 0.4em\relax Springer, 2014, pp. 818--833.

\bibitem{yosinski2014transferable}
J.~Yosinski, J.~Clune, Y.~Bengio, and H.~Lipson, ``How transferable are features in deep neural networks?'' \emph{Advances in neural information processing systems}, vol.~27, 2014.

\bibitem{vischer2021lottery}
M.~A. Vischer, R.~T. Lange, and H.~Sprekeler, ``On lottery tickets and minimal task representations in deep reinforcement learning,'' \emph{arXiv preprint arXiv:2105.01648}, 2021.

\bibitem{peebles2023scalable}
W.~Peebles and S.~Xie, ``Scalable diffusion models with transformers,'' in \emph{Proceedings of the IEEE/CVF international conference on computer vision}, 2023, pp. 4195--4205.

\bibitem{mjlab}
K.~Zakka, Q.~Liao, B.~Yi, L.~L. Lay, K.~Sreenath, and P.~Abbeel, ``mjlab: A lightweight framework for gpu-accelerated robot learning,'' \emph{arXiv preprint arXiv:2601.22074}, 2026.

\bibitem{rombach2022high}
R.~Rombach, A.~Blattmann, D.~Lorenz, P.~Esser, and B.~Ommer, ``High-resolution image synthesis with latent diffusion models,'' in \emph{Proceedings of the IEEE/CVF conference on computer vision and pattern recognition}, 2022, pp. 10\,684--10\,695.

\bibitem{vaswani2017attention}
A.~Vaswani, N.~Shazeer, N.~Parmar, J.~Uszkoreit, L.~Jones, A.~N. Gomez, {\L}.~Kaiser, and I.~Polosukhin, ``Attention is all you need,'' \emph{Advances in neural information processing systems}, vol.~30, 2017.

\bibitem{x-pred}
T.~Li and K.~He, ``Back to basics: Let denoising generative models denoise,'' \emph{arXiv preprint arXiv:2511.13720}, 2025.

\bibitem{winter2009biomechanics}
D.~A. Winter, \emph{Biomechanics and motor control of human movement}.\hskip 1em plus 0.5em minus 0.4em\relax John wiley \& sons, 2009.

\bibitem{ainsworth2023git}
S.~K. Ainsworth, J.~Hayase, and S.~Srinivasa, ``Git re-basin: Merging models modulo permutation symmetries,'' \emph{arXiv preprint arXiv:2209.04836}, 2022.

\bibitem{kornblith2019similarity}
S.~Kornblith, M.~Norouzi, H.~Lee, and G.~Hinton, ``Similarity of neural network representations revisited,'' in \emph{International conference on machine learning}.\hskip 1em plus 0.5em minus 0.4em\relax PMlR, 2019, pp. 3519--3529.

\bibitem{he2015delving}
\BIBentryALTinterwordspacing
K.~He, X.~Zhang, S.~Ren, and J.~Sun, ``Delving deep into rectifiers: Surpassing human-level performance on imagenet classification,'' 2015. [Online]. Available: \url{https://arxiv.org/abs/1502.01852}
\BIBentrySTDinterwordspacing

\end{thebibliography}


\section*{APPENDIX}
\section{Data Collection}

Our dataset consists of actor network weights collected from RL training using \texttt{mjlab}, a MuJoCo-based framework built on PPO (RSL-RL) with default hyperparameters. For each motion, the actor is parameterized as a three-layer MLP with hidden dimensions $(128,128)$, and observation normalization is disabled for both the actor and critic. The critic is implemented as a separate MLP and is used solely during training; its weights are not included in the diffusion dataset and therefore have no effect on the weight generation.

Each actor network is trained for up to 4,500 iterations with 4,096 parallel environments. To capture the local structure of the converged weight manifold, we densely record checkpoints every 100 iterations after convergence and retain eight checkpoints per training seed for each motion. The complete training configuration is summarized in Table~\ref{tab:data-collect}.

\begin{table}[b]
\centering
\caption{RL policy training configuration (locomotion: walking / motion-tracking: dancing, spin kick).}
\label{tab:data-collect}
\begin{tabular}{ll}
\toprule
Setting & Value \\
\midrule
Simulation                      & mjlab (MuJoCo) \\
Robot                           & Unitree G1 \\
Control mode & \makecell[l]{Velocity-command locomotion / \\ Motion tracking} \\
RL algorithm                    & PPO (RSL-RL) \\
Parallel environments           & $4096$ \\
Max iterations                  & $4500$ \\
Observation dimension           & $99$ / $154$ \\
Actor hidden dimensions         & $(128, 128)$ \\
Action dimension                & $29$ \\
Observation normalization  & Disabled \\
Dataset size                    & $6000$ \\
\bottomrule
\end{tabular}
\end{table}

\section{Autoencoder}
\subsection{Training Details}

During autoencoder training, we observed that operating directly on the raw actor weights yields lower reconstruction error than training on normalized weights. Consequently, all experiments reported in this work use unnormalized actor weights as the autoencoder input. To improve robustness and encourage smooth latent representations, we inject Gaussian noise into both the input weights and latent variables during training. This noise augmentation is applied only during training and is disabled at inference time. The autoencoder architecture and training hyperparameters are summarized in Table~\ref{tab:ae-arch-train}.

\begin{table}[t]
  \centering
  \setlength{\tabcolsep}{6pt}
  \caption{Autoencoder configuration: architecture and training
    hyperparameters for the walking, dancing, and spin kick motions.}
  \begin{tabular}{@{}lll@{}}
    \toprule
    & Setting & Value \\
    \midrule
    \multirow{8}{*}{\rotatebox{90}{\textbf{Architecture}}}
      & Type                                    & Two-cluster linear MLP \\
      & Cluster input dims (Walking)            & $12800,\,20253$ \\
      & Cluster input dims (Dancing, Spin kick) & $19840,\,20253$ \\
      & Input dimension $L$ (Walking)           & $33053$ \\
      & Input dimension $L$ (Dancing, Spin kick)& $40093$ \\
      & Per-cluster latent dim                  & $1024$ \\
      & Flattened latent dim                    & $2048$ \\
      & Encoder / decoder depth                 & $1$ (linear) \\
    \midrule
    \multirow{9}{*}{\rotatebox{90}{\textbf{Training}}}
      & Optimizer                & AdamW \\
      & Learning rate            & $1\times10^{-3}$ (constant) \\
      & Weight decay             & $5\times10^{-3}$ \\
      & Batch size               & $512$ \\
      & Training steps           & $1000$ \\
      & Train / validation split & $90\%/10\%$ \\
      & Loss                     & MSE \\
      & Input noise              & $1\times10^{-4}$ \\
      & Latent noise             & $1\times10^{-3}$ \\
    \bottomrule
  \end{tabular}
  \label{tab:ae-arch-train}
\end{table}

\subsection{Ablation Study}
\subsubsection{Multi Cluster vs. Single Cluster}
Our approach employs a cluster-wise autoencoder, where the first-layer weights and biases form one cluster, while the second- and third-layer weights and biases form another. To assess the impact of this design, we compare the proposed two-cluster autoencoder against a single-cluster variant that jointly encodes all network weights and biases. For a fair comparison, both models use the same latent dimensionality of 2048.

The average MSE reconstruction error across the entire motion dataset is reported in Table~\ref{tab:ae-cluster}. Despite having identical latent capacity, the multi-cluster design consistently achieves lower reconstruction error. A plausible explanation is that the first-layer parameters generally have larger magnitudes than those of the deeper layers. When all parameters are encoded jointly, these differences in scale may complicate optimization and hinder accurate reconstruction. By partitioning parameters with distinct magnitude distributions into separate clusters, the cluster-wise autoencoder reduces this optimization difficulty and improves reconstruction accuracy.

\begin{table}[!hb]
\centering
\caption{Cluster-structure ablation on three motions. MSE is the per-sample mean of reconstruction error on each dataset.}
\label{tab:ae-cluster}
\begin{tabular}{llr}
\toprule
Motion & Config & MSE \\
\midrule
\multirow{2}{*}{Walking}
  & 2-cluster              & $\mathbf{1.87 \times 10^{-4}}$ \\
  & 1-cluster (n=2048)     & $1.05 \times 10^{-3}$          \\
\midrule
\multirow{2}{*}{Spin kick}
  & 2-cluster              & $\mathbf{1.07 \times 10^{-4}}$ \\
  & 1-cluster (n=2048)     & $5.53 \times 10^{-4}$          \\
\midrule
\multirow{2}{*}{Dancing}
  & 2-cluster              & $\mathbf{6.36 \times 10^{-5}}$ \\
  & 1-cluster (n=2048)     & $3.10 \times 10^{-4}$          \\
\bottomrule
\end{tabular}
\end{table}

\section{Diffusion Transformer}
\subsection{Latent Normalization}
\label{app:diff-latent}

The diffusion model operates entirely in the latent space of a pretrained autoencoder, which remains frozen throughout training. Each actor network is encoded by the cluster-wise autoencoder and flattened into a 2048-dimensional latent representation. This dimensionality is independent of the original network weight dimension, enabling a unified diffusion architecture across all motions.

Prior to diffusion training, the latent representations are standardized using per-dimension z-score normalization. The normalized latents are partitioned into fixed-length patches and processed by a Diffusion Transformer (DiT). At inference time, latent samples are generated through the reverse diffusion process, denormalized, and decoded by the frozen autoencoder to recover complete actor network weights. Detailed architecture, training, and sampling configurations are provided in Table~\ref{tab:dit-arch} 

\begin{table}[ht]
  \centering
  \setlength{\tabcolsep}{6pt}
  \caption{DiT configuration: architecture and
    training/sampling schedule.}
  \label{tab:dit-config}
  \begin{tabular}{@{}lll@{}}
    \toprule
    & Component & Value \\
    \midrule
    \multirow{12}{*}{\rotatebox{90}{\textbf{Architecture}}}
      & Input dimension              & $2048$ \\
      & Sequence length              & $128$ \\
      & Patch size                   & $16$ \\
      & Embedding dimension          & $1024$ \\
      & Transformer depth            & $8$ \\
      & Attention heads              & $8$ \\
      & Feed-forward expansion ratio & $4$ \\
      & Activation function          & GELU \\
      & Positional encoding          & Sinusoidal \\
      & Timestep conditioning        & adaLN-Zero \\
      & Dropout probability          & $0$ \\
      & Generation setting           & Unconditional \\
    \midrule
    \multirow{14}{*}{\rotatebox{90}{\textbf{Training \& Sampling}}}
      & Denoising steps            & $1000$ \\
      & Noise schedule             & Linear ($10^{-4}\!\rightarrow\!2\times10^{-2}$) \\
      & Objective                  & $x$-prediction \\
      & Loss                       & MSE \\
      & Latent normalization       & Per-dimension z-score \\
      & Optimizer                  & AdamW ($2\times10^{-4}$) \\
      & Training steps             & 100k \\
      & Batch size                 & $256$ \\
      & Learning rate schedule     & Warmup (10k) + cosine decay \\
      & Gradient clipping          & $1.0$ \\
      & Exponential moving average & $0.9999$ \\
      & Precision                  & bfloat16 mixed precision \\
      & Sampling                   & DDPM (EMA) \\
      & Inference steps            & 1000 \\
    \bottomrule
  \end{tabular}
  \label{tab:dit-arch}
\end{table}

\section{Experiments}
\subsection{DiT Memorization vs. Generalization}
\begin{figure*}[t]
  \centering
  \begin{subfigure}{0.49\textwidth}
    \centering
    \includegraphics[width=\linewidth]{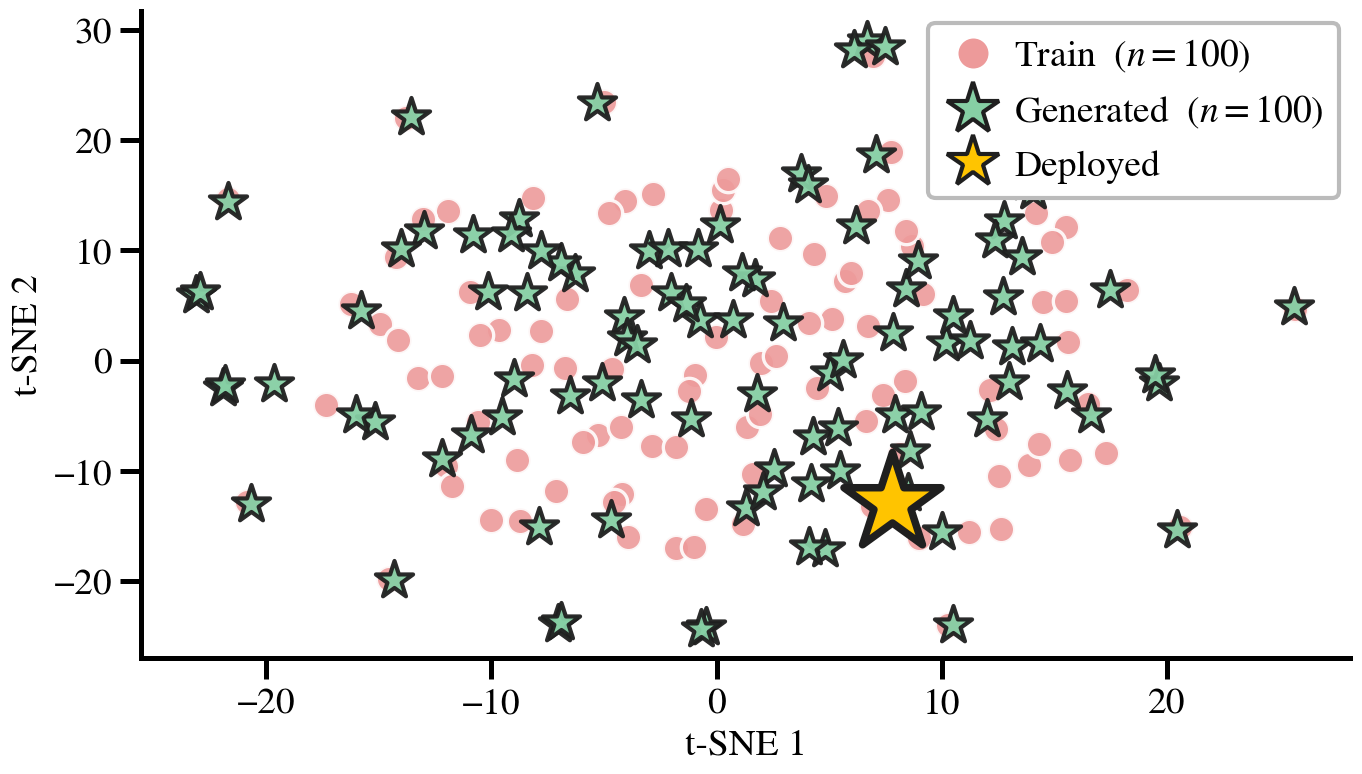}
    \caption{Walking}
    \label{fig:tsne-walking}
  \end{subfigure}
  \hfill
  \begin{subfigure}{0.49\textwidth}
    \centering
    \includegraphics[width=\linewidth]{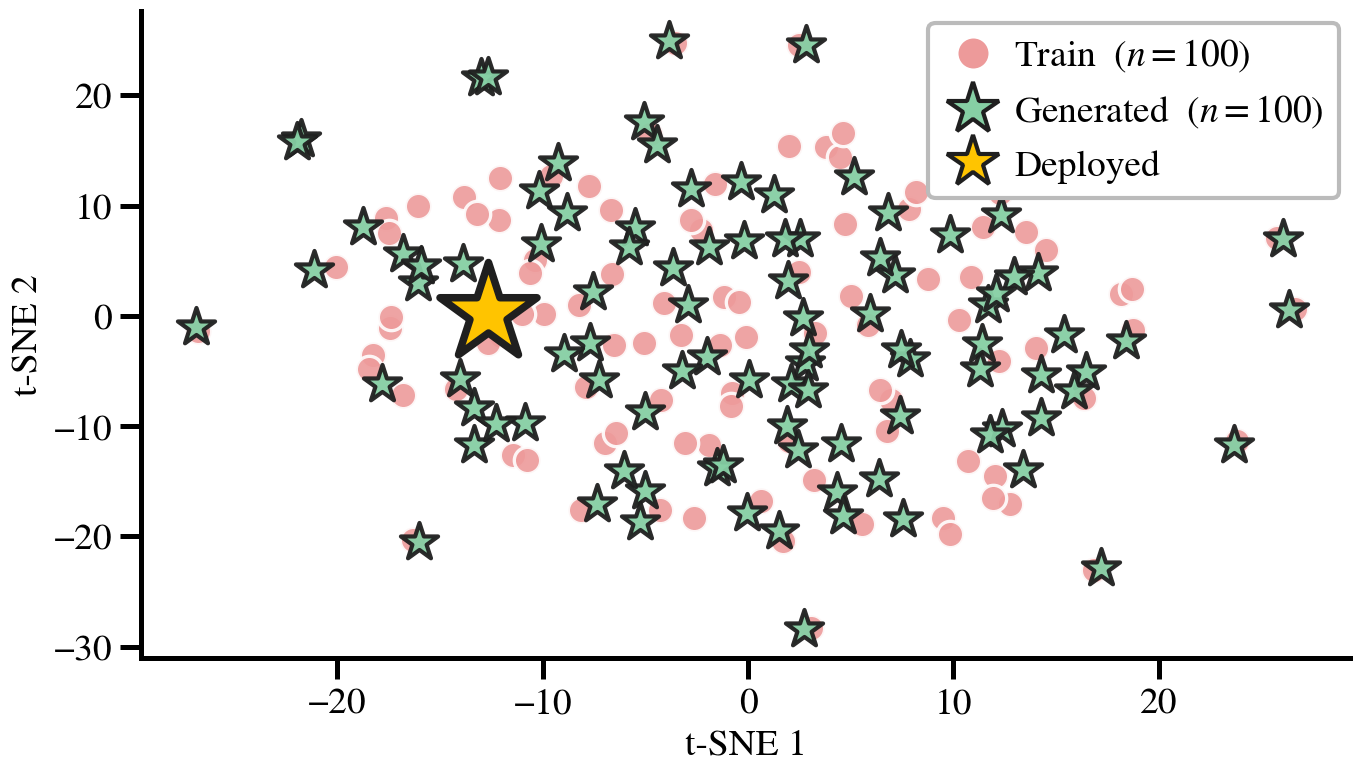}
    \caption{Dancing}
    \label{fig:tsne-dancing}
  \end{subfigure}
  \caption{\textbf{t-SNE Visualization.} Training, generated and deployed network weights for the walking and dancing motions.}
  \label{fig:tsne-additional}
\end{figure*}

We provide additional t-SNE visualizations for walking and dancing motions, each containing 100 training network weights and 100 DiT-sampled weights, as shown in Fig.~\ref{fig:tsne-additional}. The results show that the generated samples largely lie within the support of the training distribution, indicating that our method learns the data distribution.

\begin{figure}[!b]
    \centering
    \includegraphics[width=\columnwidth]{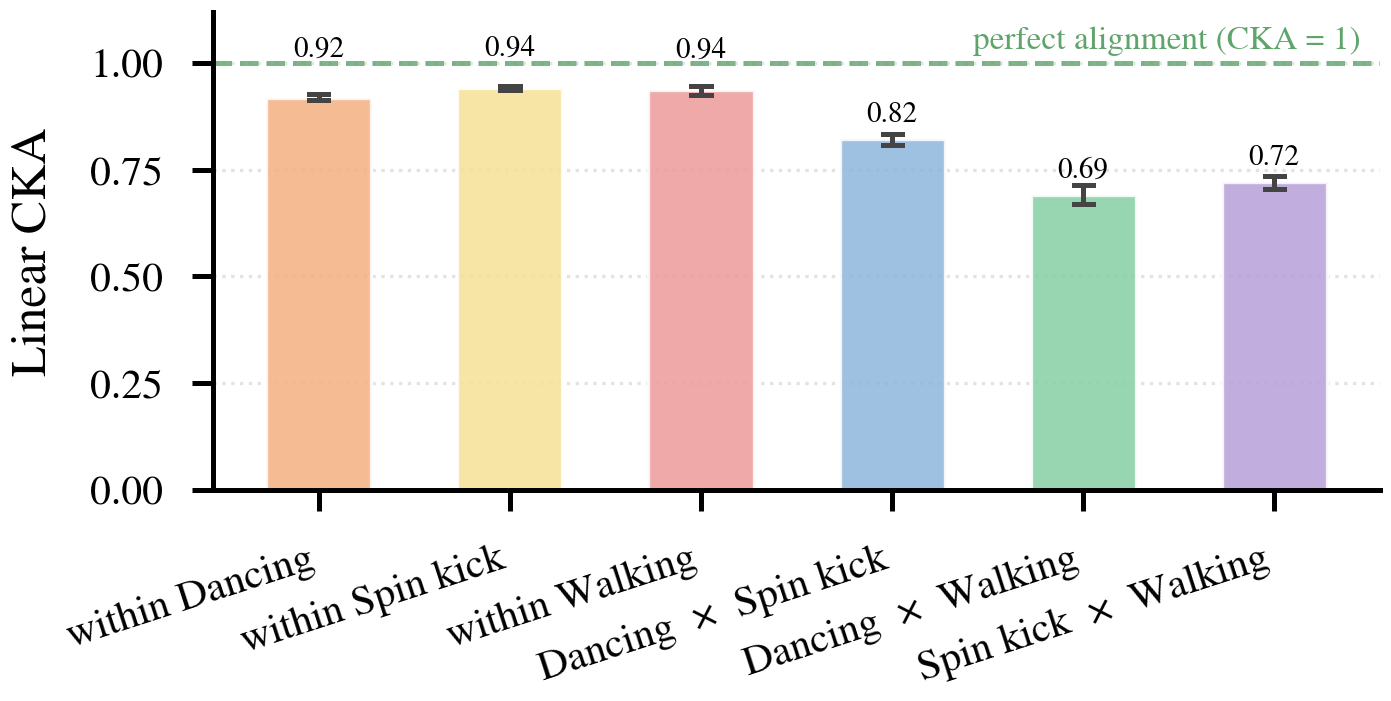}
    \caption{\textbf{Subspace similarity across motions.} Applying linear CKA to measure subspace similarity on shared inputs across motions, we observe near-perfect alignment within each motion, indicating stable knowledge extraction. Cross motion similarity also remains high, suggesting substantial shared knowledge with minor motion-specific adaptation.}
    \label{fig:cross-dataset-subspace}
\end{figure}



\section{Weight Space Visualization}
\subsection{Heatmaps for all motions}

\begin{figure*}[!ht]
  \centering
  \begin{subfigure}{\linewidth}
    \centering
    \includegraphics[width=\linewidth]{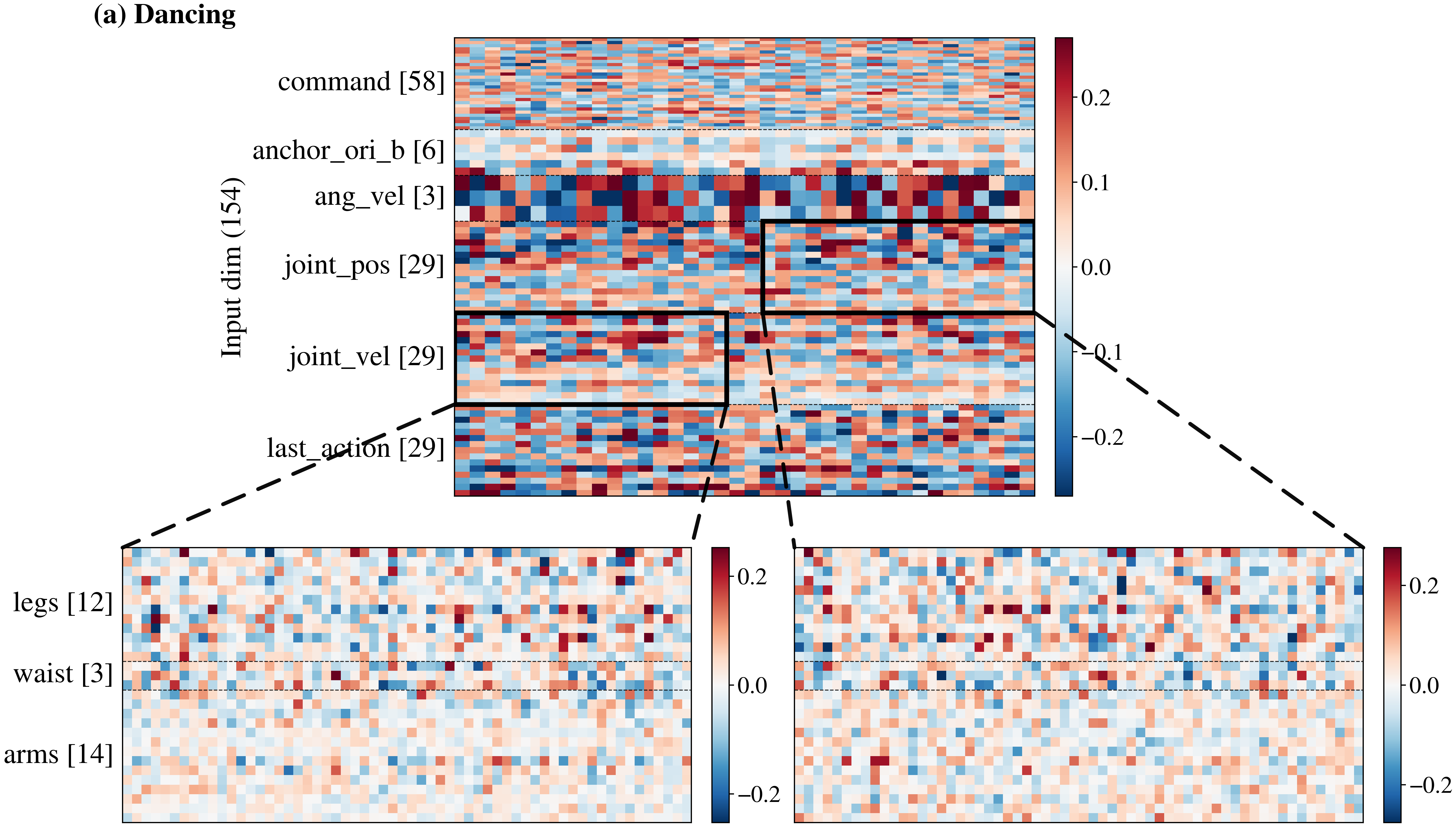}
    \label{fig:a}
  \end{subfigure}

  \begin{subfigure}{\linewidth}
    \centering
    \includegraphics[width=\linewidth]{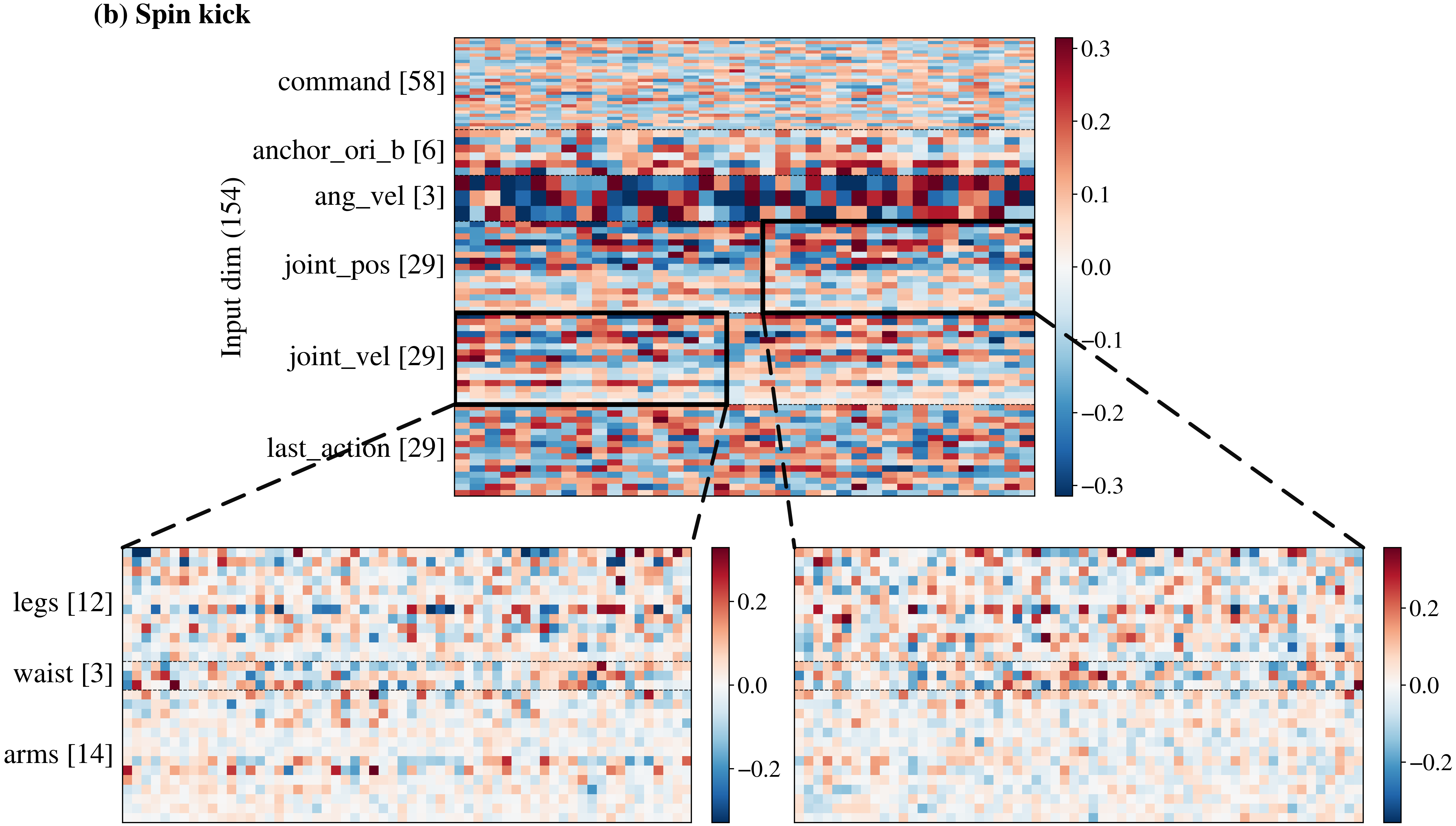}
    \label{fig:b}
  \end{subfigure}

  \caption{\textbf{First-layer heat maps}. (a) Dancing weight heat map. (b) Spin kick weight heat map. The first-layer weights consistently exhibit the structural principle across joints.}
  \label{fig:2-first}
\end{figure*}

We provide heatmap visualizations for all three layers across different motions. For walking, the first-layer heatmap is presented in the main paper in Fig.~\ref{fig:weight-region}, while here we additionally report the first-layer heatmaps for dancing and spin kick, including zoomed-in views of the \texttt{joint pos} and \texttt{joint vel} regions, as shown in Fig.~\ref{fig:2-first}. The full three-layer heatmaps across all motions are shown on \href{https://humanoid-weight-diffusion.github.io/}{the website}. Across all heatmaps, we observe the same structural principle in terms of joints described in Section~\ref{sec:sp}, which remains consistent across different motions.

\subsection{Subspace Similarity Among Motions}
The first layer contains 128 hidden nodes, each defined by a weight vector over the input dimensions. Since all motions share \texttt{joint pos} (29) and \texttt{joint vel} (29) as inputs, each hidden node corresponds to a 58-dimensional weight vector associated with these two input groups. We analyze the subspaces spanned by the resulting 128 vectors across different motions. 
Prior work has studied whether hidden nodes across different networks are identical up to permutation via canonicalization methods~\cite{ainsworth2023git}. In our datasets, we verify that the 128 vectors are \emph{not} equivalent in this sense. 
Given the structural principle in Fig.~\ref{fig:weight-region}, a natural question arises: do the first-layer subspaces show similarity across motions? To quantify subspace similarity, we adopt linear Centered Kernel Alignment (CKA)~\cite{kornblith2019similarity} to eliminate node permutation ambiguity between seeds. CKA values range from $0$ (unrelated) to $1$ (identical), and as a reference point, two independently drawn Kaiming-initialized~\cite{he2015delving} weight matrices reach a similarity of around $0.5$.

As shown in Fig.~\ref{fig:cross-dataset-subspace}, within motion seed pairs achieve CKA of $0.92$--$0.94$, confirming that different seeds converge to nearly the same subspace within each motion. Cross motion pairs yield $0.69$--$0.82$, which is lower. This indicates that the first layer is neither motion-agnostic nor motion-specific: a substantial shared knowledge is shaped by the G1 robot’s kinematics, while each motion introduces its own modulation within this shared subspace. Combined with the consistent structural principle observed earlier, these results suggest that the first layer weights primarily extract robot embodiment knowledge from the input and then adapt to the motion objective.

\subsection{RL and IL distillation for all motions}

During distillation, a student MLP with the same architecture and activation functions is trained on the RL rollout buffer using mean-squared regression, without environment interaction or reward signals. Buffer collection details are provided in Table~\ref{tab:distill_buffers}. The student is initialized from the same weights as its corresponding teacher, ensuring an identical starting point in weight space. This procedure is repeated independently for each teacher, yielding 50 teacher--student pairs per motion. The distilled IL networks are then evaluated in closed-loop simulation and achieve performance comparable to their RL teachers in both success rate (Succ) and tracking error (TE), showing functional equivalence. Training details are reported in Table~\ref{tab:distill_setup}.

\begin{table}[t]
\centering
\caption{Distillation buffers collected from RL rollouts.}
\label{tab:distill_buffers}
\setlength{\tabcolsep}{4pt}
\renewcommand{\arraystretch}{1.15}
\footnotesize
\begin{tabular}{@{}lccc@{}}
\toprule
Buffer & $N_{\mathrm{env}}\!\times\!T\!\times\!R$ & Samples & Obs \\
\midrule
Walking   & $1024\!\times\!500\!\times\!2$ & 1{,}024{,}000 & \phantom{0}99 \\
Dancing   & $512\!\times\!892\!\times\!2$   &   913{,}408 & 154 \\
Spin kick & $512\!\times\!232\!\times\!8$   &   950{,}272 & 154 \\
\bottomrule
\end{tabular}
\\[2pt]
\begin{minipage}{\columnwidth}
\footnotesize
$N_{\mathrm{env}}$ is the number of parallel environments, $T$ the episode length in frames, and $R$ the number of episodes collected per environment.
\end{minipage}
\end{table}

\begin{table}[t]
\centering
\caption{Distillation configuration. Settings are shared across all three
motions}
\label{tab:distill_setup}
\begin{tabular}{ll}
\toprule
Setting & Value \\
\midrule
Objective            & MSE \\
Optimizer            & AdamW \\
Learning rate        & $3\times10^{-4} \rightarrow 1\times10^{-5}$ (cosine) \\
Weight decay         & $1\times10^{-5}$ \\
Batch size           & 4096 \\
Epochs               & 300 \\
Train\,/\,val split  & 90\,/\,10 (seed 42) \\
\bottomrule
\end{tabular}
\end{table}

Having successfully confirmed their equivalence functionally, we provide heatmap visualizations for all motions across the three network layers to examine the differences between RL-trained and IL-trained networks. In addition to the walking results presented in the main paper, Fig.~\ref{fig:bcrl-first-layer-a} and Fig.~\ref{fig:bcrl-first-layer-b} present first-layer heatmaps for dancing and spin kick. Zoomed-in views of the \texttt{joint pos} and \texttt{joint vel} regions further highlight that the kinematic prior is consistent with those observed in walking. Complete visualizations covering all motions and all three layers are provided on  \href{https://humanoid-weight-diffusion.github.io/}{the website}. 

\subsection{Details of domain randomization in RL algorithm}
We define domain randomization (DR) as the randomization of physical and sensing parameters of the simulated system,  quantities that are fixed
properties of a given robot instance rather than part of its state. Under
this definition, our environments randomize three components: the sliding
friction of the foot-ground contact, the center-of-mass offset of the torso
link, and a constant per-joint bias on the measured joint positions. These
are summarized in Table~\ref{tab:dr-info}. Each parameter is sampled once
per parallel environment when the simulation is instantiated and held fixed
for the duration of training, so that at any gradient step the batch spans
the full parameter distribution while each individual environment remains
dynamically consistent within an episode. Training without DR disables these components, and all other aspects of the environment are unchanged.


\section{Input Meaning}

To aid the analysis of weight-space distributions, we provide detailed definitions of the actor network inputs and outputs for each motion. The input dimensions of the walking motion are described in Table~\ref{tab:walking-input}, whereas those of the dancing and spin kick (motion-tracking tasks) are summarized in Table~\ref{tab:motion-tracking-input}. For all motions, the network outputs correspond to the 29 joints of the G1 robot, whose definitions are listed in Table~\ref{tab:output} and Table~\ref{tab:joints}.
\begin{figure*}[t]
  \centering
  \includegraphics[width=\linewidth]{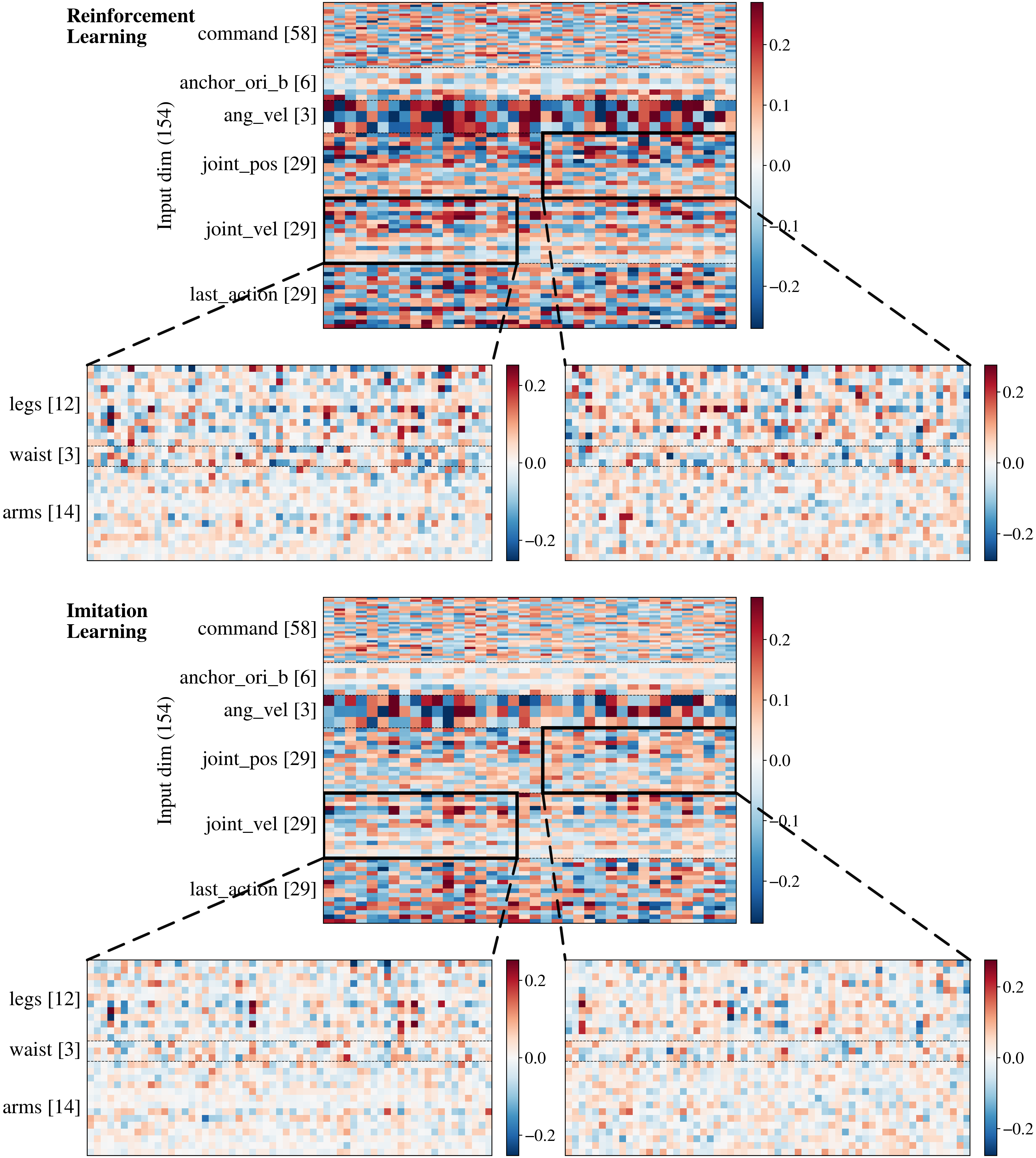}
  \renewcommand{\thefigure}{\arabic{figure}a}%
  \caption{\textbf{First-layer weight comparison between RL and IL for dancing.}}
  \label{fig:bcrl-first-layer-a}
\end{figure*}

\begin{figure*}[tp]
  \centering
  \addtocounter{figure}{-1}%
  \renewcommand{\thefigure}{\arabic{figure}b}%
  \includegraphics[width=\linewidth]{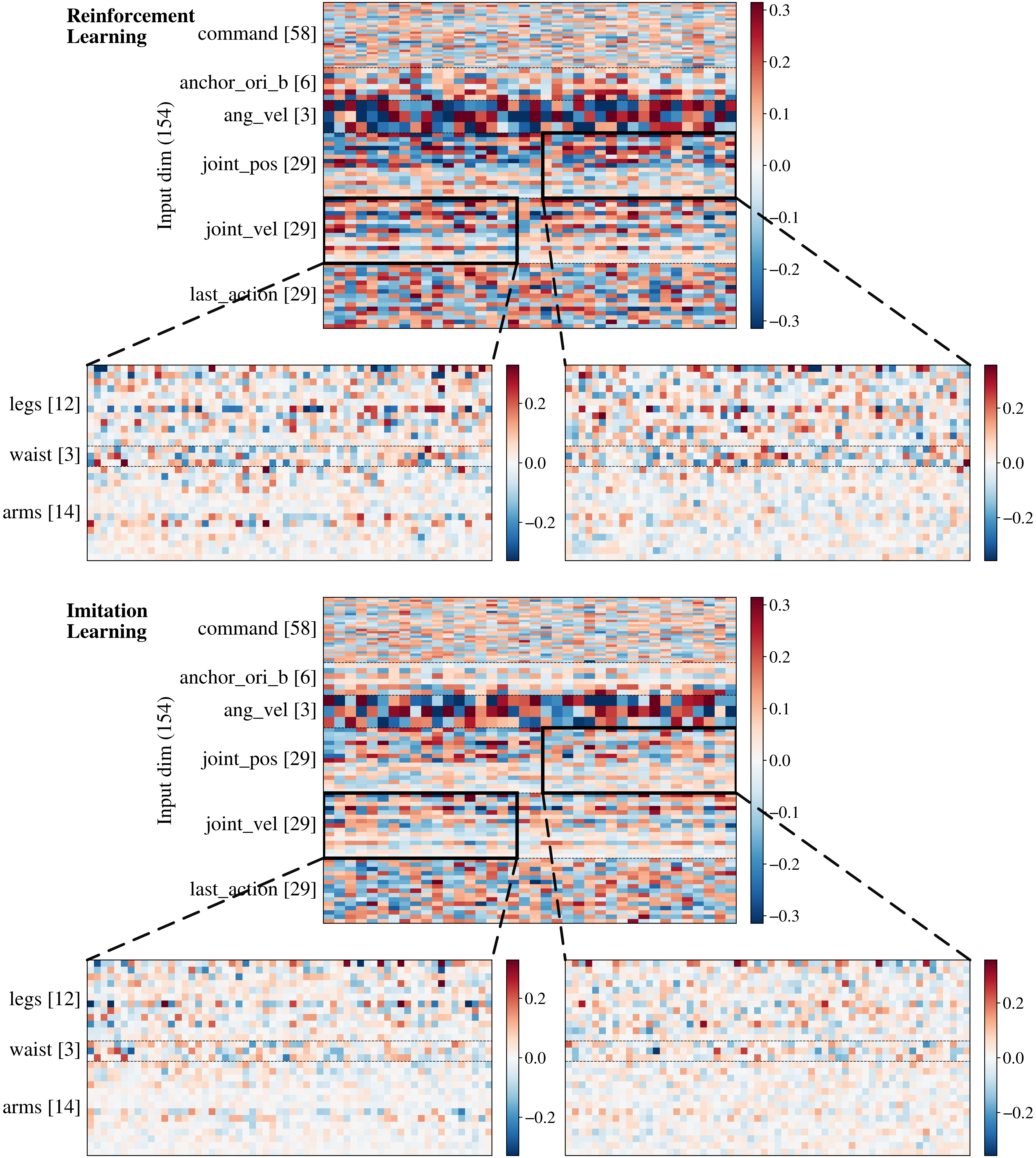}
  \caption{\textbf{First-layer weight comparison between RL and IL for spin kick.}}
  \label{fig:bcrl-first-layer-b}
\end{figure*}

\begin{table*}[!t]
\centering
\footnotesize

\independenttablecaption
  {Domain randomization terms}
  {tab:dr-info}

\setlength{\tabcolsep}{6pt}
\renewcommand{\arraystretch}{1.05}

\begin{tabular}{@{}llccl@{}}
\toprule
\textbf{Term} & \textbf{Randomized quantity} &
\textbf{Walking} & \textbf{Dancing} & \textbf{Sampling} \\
\midrule
Foot friction
  & Sliding friction coeff.\ of foot geoms
  & $\mathcal{U}(0.3,\,1.2)$
  & $\mathcal{U}(0.3,\,1.2)$
  & per env \\
Torso CoM offset
  & CoM position of torso link, per axis
  & $\pm(2.5,2.5,3.0)$\,cm
  & $\pm(2.5,5.0,5.0)$\,cm
  & per env, per axis \\
Encoder bias
  & Constant offset on measured joint pos.
  & $\pm 0.015$\,rad
  & $\pm 0.010$\,rad
  & per env, per joint \\
\bottomrule
\end{tabular}

\vspace{2cm}


\begin{minipage}[t]{0.48\textwidth}
\vspace{0pt}
\centering

\independenttablecaption
  {\textbf{Walking} actor observation, 99-d, in concatenation order.
   Joint index $k$ refers to Table~\ref{tab:joints}.}
  {tab:walking-input}

\renewcommand{\arraystretch}{0.96}

\begin{tabularx}{\linewidth}{@{}p{0.42\linewidth}X@{}}
\toprule
Block (dim) --- meaning & Per-dimension content \\
\midrule

\texttt{base\_lin\_vel} (3) --- base linear velocity (IMU)
  & $[0]\,v_x,\ [1]\,v_y,\ [2]\,v_z$ in the base frame \\

\texttt{base\_ang\_vel} (3) --- base angular velocity (IMU)
  & $[0]\,\omega_x,\ [1]\,\omega_y,\ [2]\,\omega_z$ \\

\texttt{projected\_gravity} (3) --- gravity direction
  & unit gravity vector projected into the base frame \\

\texttt{joint\_pos} (29) --- joint angles rel.\ default
  & dim $k=q_k-q_k^{\mathrm{default}}$, $k=0\dots28$ \\

\texttt{joint\_vel} (29) --- joint angular velocities
  & dim $k=\dot q_k$ \\

\texttt{last\_action} (29) --- previous action
  & dim $k$ = previous raw action of joint $k$ \\

\texttt{command} (3) --- velocity command (twist)
  & $[0]\,v_x^{*},\ [1]\,v_y^{*},\ [2]\,\omega_z^{*}$
    (target base twist) \\

\bottomrule
\end{tabularx}

\vspace{2cm}

\independenttablecaption
  {\textbf{Dancing} and \textbf{Spin kick} actor observation, 154-d,
   in concatenation order. Joint index $k$ refers to
   Table~\ref{tab:joints}.}
  {tab:motion-tracking-input}

\renewcommand{\arraystretch}{0.96}

\begin{tabularx}{\linewidth}{@{}p{0.42\linewidth}X@{}}
\toprule
Block (dim) --- meaning & Per-dimension content \\
\midrule

\texttt{command} (58) --- reference motion target
  & $[0\!:\!29]$ target joint angles $q_k^{*}$;\
    $[29\!:\!58]$ target joint velocities $\dot q_k^{*}$
    (same joint order) \\

\texttt{anchor\_ori\_b} (6) --- anchor relative orientation (6D rot.)
  & first two columns of the rotation matrix of the reference anchor frame
    w.r.t.\ the robot anchor frame:
    $[0\!:\!3]$ col.\ 1, $[3\!:\!6]$ col.\ 2 \\

\texttt{base\_ang\_vel} (3) --- base angular velocity (IMU)
  & $[0]\,\omega_x,\ [1]\,\omega_y,\ [2]\,\omega_z$ \\

\texttt{joint\_pos} (29) --- joint angles rel.\ default
  & dim $k=q_k-q_k^{\mathrm{default}}$ \\

\texttt{joint\_vel} (29) --- joint angular velocities
  & dim $k=\dot q_k$ \\

\texttt{last\_action} (29) --- previous action
  & dim $k$ = previous raw action of joint $k$ \\

\bottomrule
\end{tabularx}

\end{minipage}
\hfill
\begin{minipage}[t]{0.48\textwidth}
\vspace{0pt}
\centering

\independenttablecaption
  {Actor network output, 29-d, identical across all motions.
   Joint index $k$ refers to Table~\ref{tab:joints}.}
  {tab:output}

\renewcommand{\arraystretch}{0.96}

\begin{tabularx}{\linewidth}{@{}p{0.42\linewidth}X@{}}
\toprule
Block (dim) --- meaning & Per-dimension content \\
\midrule

\texttt{action} (29) --- joint position targets
  & dim $k$ is $a_k$, mapped to the target angle
    $q_k^{\mathrm{tgt}}
    = q_k^{\mathrm{default}}
    + \mathrm{scale}\cdot a_k$
    (\texttt{G1\_ACTION\_SCALE}),
    tracked by the joint PD controller \\

\bottomrule
\end{tabularx}

\vspace{2cm}

\independenttablecaption
  {The 29 actuated joints of the Unitree G1 in model order,
   corresponding to the ordering of the \texttt{joint pos},
   \texttt{joint vel}, and \texttt{last action} observations and
   the 29-dimensional action output. Joints are grouped by body
   region and ordered from left to right and proximal to distal.}
  {tab:joints}

\setlength{\tabcolsep}{4.5pt}
\renewcommand{\arraystretch}{0.90}

\begin{tabular}{@{}lllc@{}}
\toprule
Region & Side & Joint & Range [rad] \\
\midrule

\multirow{12}{*}{legs}
  & \multirow{6}{*}{left}
          & \texttt{hip\_pitch}      & $[-2.53,\ 2.88]$ \\
  & & \texttt{hip\_roll}             & $[-0.52,\ 2.97]$ \\
  & & \texttt{hip\_yaw}              & $[-2.76,\ 2.76]$ \\
  & & \texttt{knee}                  & $[-0.09,\ 2.88]$ \\
  & & \texttt{ankle\_pitch}          & $[-0.87,\ 0.52]$ \\
  & & \texttt{ankle\_roll}           & $[-0.26,\ 0.26]$ \\

\cmidrule(l){2-4}

  & \multirow{6}{*}{right}
          & \texttt{hip\_pitch}      & $[-2.53,\ 2.88]$ \\
  & & \texttt{hip\_roll}             & $[-2.97,\ 0.52]$ \\
  & & \texttt{hip\_yaw}              & $[-2.76,\ 2.76]$ \\
  & & \texttt{knee}                  & $[-0.09,\ 2.88]$ \\
  & & \texttt{ankle\_pitch}          & $[-0.87,\ 0.52]$ \\
  & & \texttt{ankle\_roll}           & $[-0.26,\ 0.26]$ \\

\midrule

\multirow{3}{*}{waist}
  & \multirow{3}{*}{--}
          & \texttt{yaw}             & $[-2.62,\ 2.62]$ \\
  & & \texttt{roll}                  & $[-0.52,\ 0.52]$ \\
  & & \texttt{pitch}                 & $[-0.52,\ 0.52]$ \\

\midrule

\multirow{14}{*}{arms}
  & \multirow{7}{*}{left}
          & \texttt{shoulder\_pitch} & $[-3.09,\ 2.67]$ \\
  & & \texttt{shoulder\_roll}        & $[-1.59,\ 2.25]$ \\
  & & \texttt{shoulder\_yaw}         & $[-2.62,\ 2.62]$ \\
  & & \texttt{elbow}                 & $[-1.05,\ 2.09]$ \\
  & & \texttt{wrist\_roll}           & $[-1.97,\ 1.97]$ \\
  & & \texttt{wrist\_pitch}          & $[-1.61,\ 1.61]$ \\
  & & \texttt{wrist\_yaw}            & $[-1.61,\ 1.61]$ \\

\cmidrule(l){2-4}

  & \multirow{7}{*}{right}
          & \texttt{shoulder\_pitch} & $[-3.09,\ 2.67]$ \\
  & & \texttt{shoulder\_roll}        & $[-2.25,\ 1.59]$ \\
  & & \texttt{shoulder\_yaw}         & $[-2.62,\ 2.62]$ \\
  & & \texttt{elbow}                 & $[-1.05,\ 2.09]$ \\
  & & \texttt{wrist\_roll}           & $[-1.97,\ 1.97]$ \\
  & & \texttt{wrist\_pitch}          & $[-1.61,\ 1.61]$ \\
  & & \texttt{wrist\_yaw}            & $[-1.61,\ 1.61]$ \\

\bottomrule
\end{tabular}

\end{minipage}

\end{table*}

\end{document}